\pdfoutput=1

\documentclass[11pt]{article}

\usepackage{acl}

\usepackage{times}
\usepackage{latexsym}
\usepackage{graphicx}
\usepackage{booktabs}
\usepackage{multirow}
\usepackage{amsmath}
\usepackage{amssymb}
\usepackage{subcaption}
\usepackage{enumitem}
\usepackage{comment}

\usepackage{pifont}  
\renewcommand{\checkmark}{\ding{52}}
\newcommand{\crossmark}{\ding{55}}

\newcommand{\colorcheckmark}{\textcolor[HTML]{00D32A}{\checkmark}}
\newcommand{\colorcrossmark}{\textcolor[HTML]{D3003F}{\crossmark}}

\newcommand{\checkedbox}{$\rlap{\colorcheckmark}\square$~}
\newcommand{\checkbox}{$\rlap{\hphantom{\checkmark}}\square$~}
\newcommand{\crossedbox}{$\rlap{\colorcrossmark}\square$~}

\usepackage[T1]{fontenc}

\usepackage[utf8]{inputenc}

\usepackage{microtype}

\title{Possible Stories: Evaluating Situated Commonsense Reasoning \\ under Multiple Possible Scenarios}

\author{Mana Ashida\thanks{\hphantom{x}Work done while at Tokyo Metropolitan University.}\\
  Yahoo Japan Corporation\\
  \texttt{maashida@yahoo-corp.jp} \\\And
  Saku Sugawara \\
  National Institute of Informatics \\
  \texttt{saku@nii.ac.jp} \\}

\begin{document}
\maketitle
\begin{abstract}
The possible consequences for the same context may vary depending on the situation we refer to. 
However, current studies in natural language processing do not focus on situated commonsense reasoning under multiple possible scenarios.
This study frames this task by asking multiple questions with the same set of possible endings as candidate answers, given a short story text.
Our resulting dataset, Possible Stories, consists of more than 4.5K questions over 1.3K story texts in English.
We discover that even current strong pretrained language models struggle to answer the questions consistently, highlighting that the highest accuracy in an unsupervised setting (60.2\%) is far behind human accuracy (92.5\%).
Through a comparison with existing datasets, we observe that the questions in our dataset contain minimal annotation artifacts in the answer options.
In addition, our dataset includes examples that require counterfactual reasoning, as well as those requiring readers' reactions and fictional information, suggesting that our dataset can serve as a challenging testbed for future studies on situated commonsense reasoning.
\end{abstract}

\section{Introduction}
Commonsense reasoning, inclusive of counterfactual, abductive, and monotonic reasoning, is a core element of language understanding. 
Researchers are interested in whether these abilities can be learned in systems, and several benchmarks have been proposed to investigate machine commonsense reasoning~\citep{huang-etal-2019-cosmos, sap-etal-2019-social,aggarwal-etal-2021-explanations,saha-etal-2021-explagraphs}.
Recent pretrained models have shown competitive results~\citep{khashabi-etal-2020-unifiedqa, Lourie2021}.

\begin{figure}[t]
\centering
    \includegraphics[width=\columnwidth]{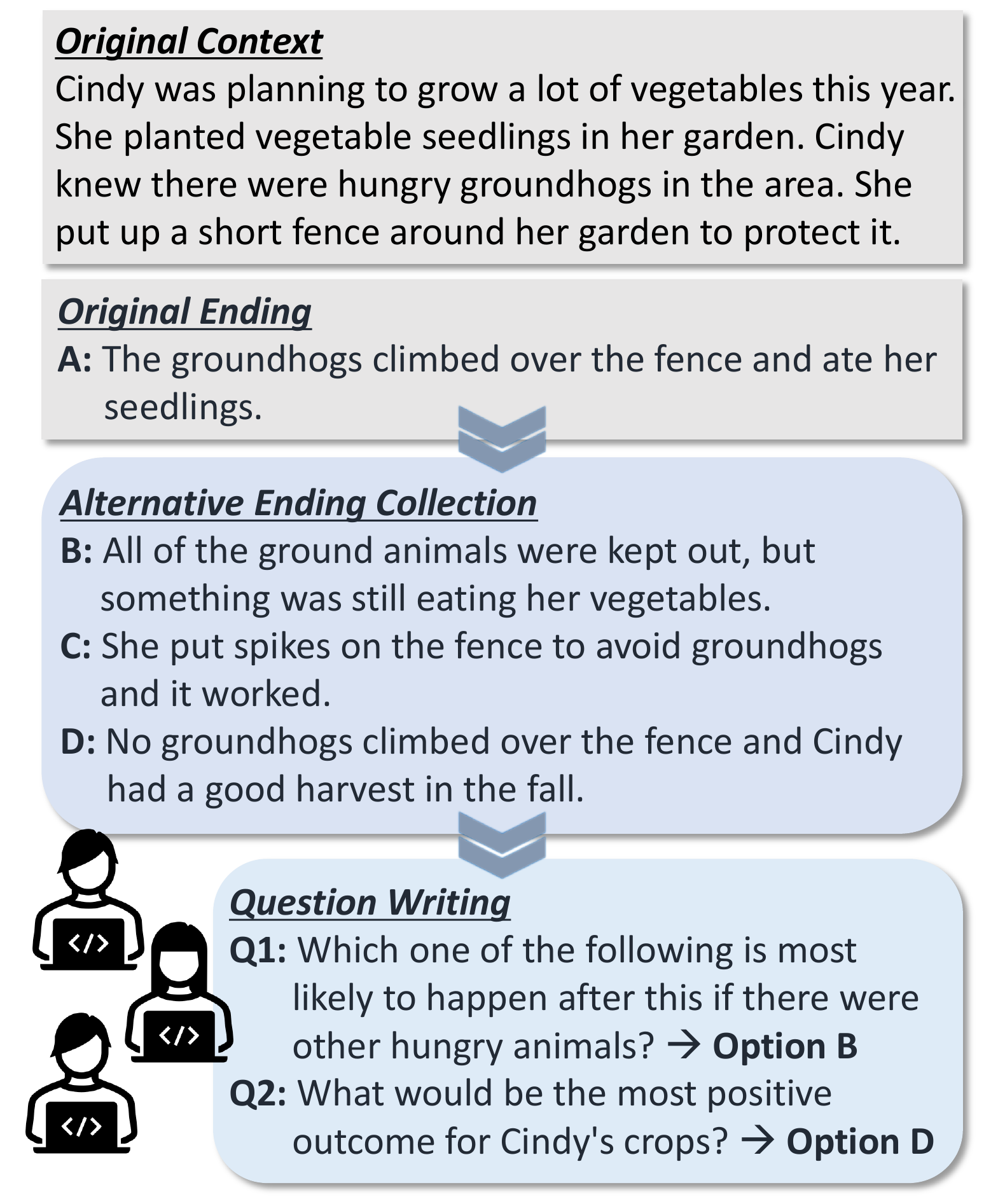}
    \caption{
    Overview of Possible Stories and its creation process.
    We ask crowdworkers to produce alternative endings given a story text, and then write multiple-choice questions that have a single correct answer among the original and three collected endings.
    }
    \label{fig:running_example}
\end{figure}

Commonsense reasoning has often been framed as a task to infer whether candidate answers are plausible, such as in the multiple-choice format \cite{talmor-etal-2019-commonsenseqa,sakaguchi-etal-2020-winogrande}.
The difference between plausible and implausible answers is expected to be salient enough that it can be established as a classification task.
However, when making day-to-day decisions, people consider several plausible choices, rather than clearly plausible and implausible ones, depending on one's situation and method of thinking.
However, the tasks concerning conditions \textit{under multiple plausible scenarios} are few, and their domains are limited to, for example, factual information that differs according to place and time \cite{zhang-choi-2021-situatedqa} or human behaviors that are either normative or divergent \cite{emelin-etal-2021-moral}.
Another example is the natural language inference or commonsense reasoning task that considers variations in human opinions~\citep{zhang-etal-2017-ordinal, chen-etal-2020-uncertain}, which allows for the differences in annotations due to one's mentality~\cite{pavlick-kwiatkowski-2019-inherent, meissner-etal-2021-embracing}. 
Our aim here is to interrogate these types of situated reasoning in more comprehensive settings, such as in story texts.

To assess the possible extent of situated reasoning in machines, we introduce \textit{Possible Stories}, a benchmark consisting of 4,533 multiple-choice questions over 1,313 passages in English, to evaluate commonsense reasoning over multiple possible alternatives for single passages.
Figure~\ref{fig:running_example} shows an example.
Given a story text, we aggregate alternative endings and multiple-choice questions that contain information such that they guide the determination of the most likely ending.
By design, machines cannot rely only on answer options but also have to understand the condition implied by each question to answer correctly because all options are expected to be possible.
This dataset creation procedure tackles the known issue of annotation artifacts~\cite{gururangan-etal-2018-annotation} in crowdsourced datasets by using alternative endings, instead of \textit{right} and \textit{wrong} endings, and by compiling the endings and questions from multiple crowdworkers.

We evaluate strong pretrained language models and heuristic methods on our dataset and observe that in an unsupervised setting, even the strongest model \cite[DeBERTa large v3;][]{he2021debertav3} underperforms compared to humans by approximately 30\% accuracy and more than 50\% consistency score (i.e., passage-wise accuracy).
Our analysis using input ablation and statistical significance tests highlights that the annotation artifacts contained in the answer options of our dataset questions are much fewer than those in existing multiple-choice datasets such as RACE \cite{lai-etal-2017-race} and CosmosQA \cite{huang-etal-2019-cosmos}.
Reasoning-type annotation shows that more than 60\% of our dataset questions require counterfactual reasoning, as well as an understanding of characters' motivations and reactions, readers' perceptions, and fictional information.

Our contributions are summarized as follows:\footnote{The details of our data collection and final outcome including all collected story endings are available at \url{https://github.com/nii-cl/possible-stories}.} 
\begin{itemize}[leftmargin=1em]
    \item We propose a situated commonsense reasoning task and create a multiple-choice question answering (QA) dataset using plausible story endings, together with questions as multiple conditions where one of the endings becomes the most plausible.
    \item We discover that current strong pretrained language models struggle to solve our task when training data are unavailable, indicating that there is room for future improvement on situated commonsense reasoning.
    \item We show that our dataset contains minimal annotation artifacts in the answer options and has many challenging questions that require counterfactual reasoning and an understanding of characters' motivations and reactions, readers' perceptions, and fictional information.
\end{itemize}
 
\section{Background and Related Works}
Our work is motivated by recent efforts to create evaluation frameworks for commonsense reasoning situated in extra-linguistic contexts.

\paragraph{Benchmarks for Commonsense Reasoning}
Many commonsense reasoning resources have been proposed that target reading comprehension \cite{huang-etal-2019-cosmos}, cloze tests regarding story endings \cite{zellers-etal-2019-hellaswag} or in-between events \cite{Bhagavatula2020Abductive}, and inferences on social interactions \cite{sap-etal-2019-social}. 
\citet{mostafazadeh-etal-2016-corpus} propose a task similar to ours, but it differs in that ours has four possible ending options, rather than a plausible and implausible completion pair.

\paragraph{Benchmarks for Counterfactual Reasoning}
Researchers have coined the term \textit{counterfactual reasoning} to refer to the property of reasoning over hypothetical events and have proposed benchmarks to evaluate the counterfactual reasoning ability of machines.
\citet{tandon-etal-2019-wiqa} collect questions that explicitly ask \textit{what if}, based on procedural texts.
\citet{qin-etal-2019-counterfactual} propose a task of generating a counterfactual story ending that is minimally edited from the original ending, given modified events in the context. 
Our data creation process is similar in terms of using an existing story and modifying a segment; however, we ask crowdworkers to change the segment more drastically, yielding diverse story endings.

\paragraph{Evaluation of Understanding of Situations}
Reasoning over multiple possibilities, depending on the situation, can be regarded as \textit{situated reasoning}. 
Recent studies have attempted to integrate situational information into the context used in downstream tasks, such as question answering on factual information~\cite{zhang-choi-2021-situatedqa} and consequence or normative action generation given real-world social settings~\cite{emelin-etal-2021-moral}.
Story Commonsense~\citep{rashkin-etal-2018-modeling} provides an annotated dataset of motivation and emotional reactions. 
\cite{forbes-etal-2020-social} collect general rules of thumb about actions.
The range of situations that we consider goes beyond facts and normative settings, aiming to consider readers' beliefs, causality, and characters' emotions.


\paragraph{Probing of Language Models}
The use of contrastive examples to probe language models' knowledge and inductive biases is an active area of research. 
This line of research typically uses pairs of sentences with minimum differences~\citep{marvin-linzen-2018-targeted, li-etal-2020-unqovering, warstadt-etal-2020-blimp-benchmark}, contrastive sets to identify the model's decision boundary~\cite{gardner-etal-2020-evaluating}, or adversarial examples~\citep{jia-liang-2017-adversarial} to identify the segments that contribute to changing model behaviors. 
By contrast, we use multiple plausible choices for a single passage to study what causes models to assign higher probabilities to certain choices.

\section{Task Description}

\paragraph{Motivation}
In Story Cloze Test, \citet{mostafazadeh-etal-2016-corpus} use \textit{right} and \textit{wrong} endings to evaluate machines' story understanding, assuming that the right ending can be regarded as an entailing hypothesis in a textual entailment framework and the wrong ending as a contradicting hypothesis.
During data collection, the workers are instructed to produce endings that are realistic and sensible for right endings, and wrong endings are chosen from those that are rated lower than right endings in terms of meaningfulness and coherence.
Consequently, their task is created to have clear right and wrong endings.
However, in reality, there is an infinite number of possibilities of clearly plausible endings.
By creating informative questions positing situations rather than questions asking about relative plausibility without any conditions, we aim to test machines' story understanding in multiple scenarios that provide additional information that can discriminate one plausible ending from other possible endings.

\paragraph{Task Formulation}
We formulate the task as a multiple-choice question with a passage and answer options, where the answer options depict possible endings of the passage. 
Given a story $s$, the task is to determine the most plausible story ending among the four possible endings $E=(e_1, \ldots, e_4)$ under the condition $c$ that is implied by a question.
To further evaluate the models' understanding of situations, we also define the task of predicting the most plausible outcome for multiple conditions.
Given $s$, the task is to determine the most plausible story ending among $E$ for each of the multiple conditions $C=(c_1, \ldots, c_4)$ that are implied by multiple questions. 
When the models capture all the relationships between conditions and plausible endings correctly, we assume that the models reason over a finite number of possible consequences and the relationships among them. 
We call this \textit{consistency}, which reports the percentage of a model's outputs that are correct for all questions referring to a unique context.
This evaluation is inspired by the study of contrastive sets~\citep{gardner-etal-2020-evaluating}.

\section{The Possible Stories Dataset}

\paragraph{Context Passages} 
To collect story texts, we use ROCStories \cite{mostafazadeh-etal-2016-corpus}, a corpus of five-sentence stories.
The first to fourth sentences describe the context, and the final sentence, the ending, concludes the story.
We choose ROCStories because each of its stories has a clear beginning and ending, while being generic enough to come up with different endings.
The details on our story selection criteria are provided in Appendix~\ref{sec:story_selection}.

The following tasks are carried out by the crowdworkers in Amazon Mechanical Turk (MTurk) who perform above certain levels during our worker recruitment phase, which is designed to be fairer than the conventional qualifications used in MTurk.
The details of the worker recruitment are provided in Appendix~\ref{app:worker}.
The instructions and task interface presented to the workers are also provided in Appendix~\ref{app:annotation-instructions}.

\subsection{Writing Tasks}

\paragraph{Ending Writing} 
We first ask workers to create two alternative endings given a story with the original ending.
The participants are encouraged to be as creative as possible so that possible yet unrealistic story endings can also be elicited. 
We collect six to eight alternative endings by asking three or four workers to produce two endings per passage.

\paragraph{Selection of Ending Options}
Having collected six to eight alternative endings, we need to decide which three options to use in our questions, in addition to the original ending.

Depending on how they are chosen, there may be differences in the difficulty of the generated questions.
For example, if the endings are similar to each other, it will be difficult to create questions that have only one correct choice among four endings. 
Conversely, if the endings are completely different, the questions may be easier to create, but machines may rely solely on semantic similarity between passage and endings, without requiring commonsense reasoning.

To examine the relationship between question difficulty and the diversity of the chosen endings, we run a pilot task using ten randomly selected stories with six different sets of endings. 
The six sets are chosen based on the sum of cosine similarity calculated based on the embeddings~\citep{reimers-gurevych-2019-sentence} of all the possible combinations of endings, ranging from the set of endings that are most similar to the most diverse set.
Six sets are chosen such that the distance between the values of the sum of the cosine similarities of one set and another set is equal.
Through a validation step to identify which sets of endings enable high-quality multiple-choice questions, we decide on the set that contains the second most diverse endings among the six sets upon consideration.

\paragraph{Question Writing}
As four distinct endings are gathered per passage, we ask the workers to write questions in which only one of the four endings is the correct answer.
Because this task is more complex, we select participants via a qualification task, targeting those participants who maintain quality in the ending writing task.
It is up to the workers to decide the correct option, considering the difference in difficulty in writing questions with certain story endings.
Four questions are written per passage by two workers, two per worker, and the answers to each set of two questions are different to maintain the diversity of the correct answers.
To ensure that the distribution of the dataset is natural~\citep{bowman-dahl-2021-will,kaushik-etal-2021-efficacy} and that the questions fit the general purpose, we avoid collecting questions in an adversarial manner \cite{bartolo-etal-2020-beat}.

\subsection{Data Validation}
\label{sec:data-validation}

The goal of the validation task is to verify that there is one correct answer for each question, and that the questions do not contain any objectionable or personal content. 
Questions that do not meet these criteria are discarded.
The detailed validation results and further quality control over the collection batches are reported in Appendix~\ref{app:validation}.

\paragraph{Question-Answer Validation}
In this step, we ask workers to answer multiple-choice questions. 
The workers choose one of the four endings and four additional options (no answer, more than two possible answers, ill-formed questions, and others).
Each question is validated by three workers, and we retain questions in which the majority vote is identical to the writer's answer.

\paragraph{Content Validation}
During the validation task, we ask workers to indicate negative stereotypes or biased descriptions of certain social groups. 
We discard questions that the workers claim contain unfair descriptions.
This process prevents the perpetuation of unethical opinions in downstream tasks when this dataset is used for training models.
Some of the workers' inputs are discussed in Ethical Considerations.
We incentivize workers with a bonus of \$0.3 per completion of the free-text form.

\subsection{Dataset Statistics}

\begin{table*}[t] 
    \centering 
    \begin{tabular}{lrrrrrr}
    \toprule
    Split & \#Question & \#Passage & \#Q/P & Passage len & Question len & Option len \\\midrule
    Train & 3,404 & 984 & 3.46 & 46.1 & 13.9 & 15.4 \\
    Dev & 458 & 133 & 3.44 & 46.2 & 14.9 & 15.3 \\
    Test & 671 & 196 & 3.42 & 47.0 & 15.0 & 15.2 \\ \midrule
    Total & 4,533 & 1,313 & 3.45 & 46.3 & 14.2 & 15.3  \\\bottomrule
    \end{tabular}
    \caption{
    Statistics of Possible Stories.
    \textit{Q} and \textit{P} indicate question and passage.
    \textit{\#Q/P} indicates the average number of questions per passage.
    \textit{Len} indicates the average number of tokens.
    }
    \label{tab:dataset_stats}
\end{table*}

Our dataset, Possible Stories, has 8,885 alternative endings for 1,313 passages and 4,533 multiple-choice questions with the original ending and three alternative endings as answer options.
Table~\ref{tab:dataset_stats} presents the basic statistics for the resulting dataset.
Although the passages are shorter than those in CosmosQA (70.3) and RACE (321.9), the questions (14.2) and answer options (15.3) are quite longer than others (CosmosQA is 10.6 and 8.1 and RACE is 10.0 and 5.3), which could potentially make questions difficult \cite{nangia-etal-2021-ingredients}.

\begin{table}[t] \centering
\begin{tabular}{rrrrrr}
\toprule
& \multicolumn{5}{c}{Distinct \# of answers} \\ \cmidrule{2-6}
\#Q/P   & 1   & 2    & 3    & 4   & total     \\ \midrule
1   & 2.1 &      &      &     & 2.1  \\
2   & 1.5 & 8.4  &      &     & 9.9 \\
3   &     & 16.8 & 12.0 &     & 28.8 \\
4   &     & 18.1 & 34.7 & 6.5 & 59.3 \\ \midrule
Total & 3.6 & 43.3 & 46.7 & 6.5 & 100.0  \\ \bottomrule
\end{tabular}
\caption{Distribution (\%) of the number of questions per passage and the distinct number of correct answers.} 
\label{tab:dataset_distribution}
\end{table}

In addition, as shown in Table~\ref{tab:dataset_distribution}, more than 50\% of the contexts have questions with three or four distinct correct answer choices.
This contributes to the assessment of the models' comprehension of multiple situations using the consistency metric.

One of our main goals for constructing a benchmark is to test the models' capacity for situated commonsense reasoning over multiple scenarios as an unseen task.
Nonetheless, to ensure that it is feasible to model our task using current strong pretrained language models \cite{liu-etal-2019-inoculation}, we follow a standard approach to split the collected examples into training (75\%), dev (10\%), and test (15\%) sets.
To investigate the model generalizability, the passages do not overlap between each set.
The dev and test sets do not contain questions produced by workers who have received negative comments from other workers to ensure quality.

\section{Experiments}
\label{sec:experiments}

\subsection{Models and Settings}

For modern pretrained language models, we use BERT~\citep[base and large;][]{devlin-etal-2019-bert}, RoBERTa~\cite[base and large;][]{Liu2019RoBERTaAR}, and DeBERTa~\citep[base and large of v3;][]{he2021debertav3}.
In our standard setting (i.e., unsupervised), to adapt these models to the multiple-choice task, we fine-tune them on the RACE dataset \cite{lai-etal-2017-race}, which is a large-scale dataset of middle- and high-school English exams and has passages and questions on various topics.\footnote{We observe that models fine-tuned on CosmosQA are consistently inferior to those fine-tuned on RACE (Appendix~\ref{app:cosmosqa-finetune}).}
In the supervised setting, the models are directly trained on our training set unless mentioned otherwise.
To establish different baseline methods, we consider simple heuristics using perplexity, semantic similarity, and entailment scores.
For perplexity heuristics, we use GPT-2 \cite{radford2019language} and GPT-Neo \cite{gpt-neo} to obtain the perplexity of the inputs and consider options with the smallest perplexity as a model's prediction.
Sentence similarity uses representations obtained from the sentence transformers \citep{reimers-gurevych-2019-sentence} to compute the cosine similarity between the options and the rest of the input.
The candidate with the highest similarity score is regarded as the model prediction. 
The entailment score is calculated using RoBERTa-large fine-tuned on \textsc{MNLI}~\citep{williams-etal-2018-broad}
, and the option with the highest entailment score when taking the inputs as the premise is chosen.

\subsection{Results}

\paragraph{Human Performance}
To measure the human performance on our test set, we collect three additional labels for all questions from different crowdworkers who do not join the validation task.
We ensure that the same set of three workers answer the questions belonging to a single story.
For computing accuracy, we take the majority of the three labels to determine whether it is equal to the validated gold label.
For computing consistency, we determine whether the majority vote answers are correct for all questions in each passage (Table~\ref{tab:result}).

\begin{table}[t]
    \centering
\begin{tabular}{clrr}
\toprule
 FT & Model & Acc.  & Consist. \\ \midrule 
 
 \multirow{7}{*}{\crossmark}
 & DeBERTa-large$^*$ & \textbf{60.2} & \textbf{19.9} \\ 
 & DeBERTa-base$^*$ & 45.3 & 8.2 \\ 
 & RoBERTa-large$^*$ & 50.5 & 13.8 \\ 
 & PPL. GPT-2 large & 30.4 & 2.0 \\ 
 & PPL. GPT-Neo 2.7B & 29.5 & 2.6 \\ 
 & Semantic Sim. & 37.3 & 4.1 \\ 
 & Entailment & 23.1 & 2.0 \\ 
 \midrule
\multirow{8}{*}{\checkmark}
 & DeBERTa-large$^*$ & \textbf{92.1} & \textbf{74.7} \\ 
 & DeBERTa-large & 88.5 & 67.3 \\ 
 & DeBERTa-base & 81.5 & 51.5 \\ 
 & RoBERTa-large$^*$ & 83.5 & 55.6 \\ 
 & RoBERTa-large & 81.7 & 49.5 \\ 
 & RoBERTa-base & 72.0 & 30.6 \\ 
 & BERT-large & 62.6 & 20.4 \\ 
 & BERT-base & 57.3 & 16.3 \\ \midrule 
 & Human & 92.5 & 76.5 \\ \bottomrule 
\end{tabular}
%
\caption{
    Model and human performances on our dataset. \textit{Acc.}~and \textit{consist.}~denote accuracy (\%) and consistency (\%).
    $(^*)$ indicates that the model is fine-tuned on RACE.
    \textit{FT} indicates whether the models are fine-tuned on the training set.
    The experimental details are reported in Appendix~\ref{app:experiments-detail}.
}
\label{tab:result}
\end{table}

\paragraph{Model Performance} 
When the training set is unavailable, we observe that DeBERTa-large achieves the best performance.
Although this model is fine-tuned on RACE, which has a sufficient number of diverse training examples, the model performance is far from that of humans, showing large gaps of 29.5\% and 53.0\% in terms of accuracy and consistency, respectively.
Out of the four simple heuristics models, those using perplexity and semantic similarity perform above the chance rate of 25\%, indicating that those features, while inadequate, might be useful in finding the correct answers.
By contrast, the entailment score-based model falls short of 25\%. 
This result highlights the uniqueness of our dataset, as it shows that relying on monotonic reasoning cannot lead to a correct answer.

\paragraph{Supervised Performance}
With the training data, we observe that DeBERTa-large performs better than the other models, and it achieves the best accuracy and consistency when fine-tuned using RACE (Table~\ref{tab:result}).
These scores are very close to those of humans, which implies that the task can be feasibly performed by a strong model, given sufficient training data.
Nonetheless, it is notable that BERT-large and RoBERTa-large, which were state-of-the-art models only several years ago, show potential for improvement ($\approx$ 30\% and 10\% accuracy, respectively) compared to humans.

\begin{table}[t]\setlength{\tabcolsep}{5pt}
    \centering
    \begin{tabular}{clrrr} \toprule
    FT & Model & Full & No pas. & No ques. \\ \midrule
    \multirow{4}{*}{\crossmark} & DeBERTa-L$^*$ & 60.2 & 58.1 & 21.8  \\
     & RoBERTa-L$^*$ & 50.5 & 50.3 & 21.5 \\
     & GPT-2 large &  30.4 & 35.2 & 26.4 \\
     & Semantic Sim. & 37.3 & 47.1 & 28.8 \\ \midrule
    \multirow{3}{*}{\checkmark} & DeBERTa-L$^*$ & 92.1 & 87.0 & 31.8 \\
    & DeBERTa-L &  88.5 & 86.4 & 33.4 \\
    & BERT-L & 62.6 & 51.1 & 30.4 \\
    \bottomrule
    \end{tabular}
    \caption{
        Input ablation results (accuracy; \%).
        \textit{No pas.}~and \textit{no ques.}~indicate that the context passage and question are ablated from the input, respectively.
    }
    \label{tab:ablation}
\end{table}

\paragraph{Input Ablation}
Table~\ref{tab:ablation} presents the input ablation analysis.
When ablating the passages, we observe that pretrained language models fine-tuned on any multiple-choice dataset show lower performance than those with the full input.
Regarding the heuristics methods, we find that having the context does not significantly change the ranking of the endings.
When ablating the questions, we observe that the performance of all models decreases, which is expected because the same set of answer options has multiple questions in our dataset.

\section{Analysis}

\subsection{Human--model Performance Gap}
\label{sec:performance-gap}

\begin{table}[t] \setlength{\tabcolsep}{4pt}
    \centering
    \begin{tabular}{crrrr} \toprule
        Model & Ours & Cosmos & QuAIL & MC-adv \\ \midrule
        DeBERTa-L & 60.2 & 66.8 & 76.3 & 81.2 \\
        DeBERTa-B & 45.3 & 56.0 & 66.2 & 69.0 \\
        RoBERTa-L & 50.5 & 64.2 & 70.3 & 69.1 \\ \midrule
        Human & 92.5 & 94.0 & 100.0 & 92.0  \\
        Acc. gap & 40.5 & 31.7 & 29.1 & 18.9 \\ \bottomrule
    \end{tabular}
    \caption{
        Human performance, model performance without fine-tuning (accuracy; \%), and the gap between the human performance and the average model performance (larger values imply higher difficulty).
        Model-L and -B indicate the large and base models respectively.
    }
    \label{tab:other-dataset-result}
\end{table}

To investigate the relative difficulty of our dataset among multiple-choice QA datasets, we compare the accuracy gap between humans and models in an unsupervised setting with existing datasets, including CosmosQA (Cosmos; we report the validation result because the test labels are not available), QuAIL~\citep[][challenge set]{Rogers2020GettingCT}, and the examples provided by \citet{sugawara-etal-2022-makes} (MC-adv; multiple-choice questions that are written by crowdworkers in an adversarial manner).
We use three models (DeBERTa-large and -base and RoBERTa-large) fine-tuned on RACE.
The results in Table~\ref{tab:other-dataset-result} demonstrate that our dataset may be more challenging than the multiple-choice reading comprehension datasets we analyze, despite the simplicity of our data collection method.

\subsection{Annotation Artifacts in Answer Options}
\label{sec:annotation-artifact}

\begin{table}[t] \setlength{\tabcolsep}{4pt}
    \centering
    \begin{tabular}{crrrr} \toprule
        Dataset & Full & No pas. & No ques. & Opt. only \\ \midrule
        Ours & 88.5 & 86.4 & 33.4 & 29.1 \\
        RACE & 87.9 & 60.1 & 69.8 & 46.6 \\
        QuAIL & 81.7 & 51.8 & 58.3 & 39.6 \\
        Cosmos & 87.8 & 72.5 & 59.4 & 57.2 \\ \bottomrule
    \end{tabular}
    \caption{
        Supervised accuracy (\%) by DeBERTa-large (v3) on the input-ablation settings.
    }
    \label{tab:ablation-with-others}
\end{table}

\begin{figure*}[t]
    \centering
    \begin{subfigure}{.49\textwidth}
        \centering
        \includegraphics[width=\linewidth]{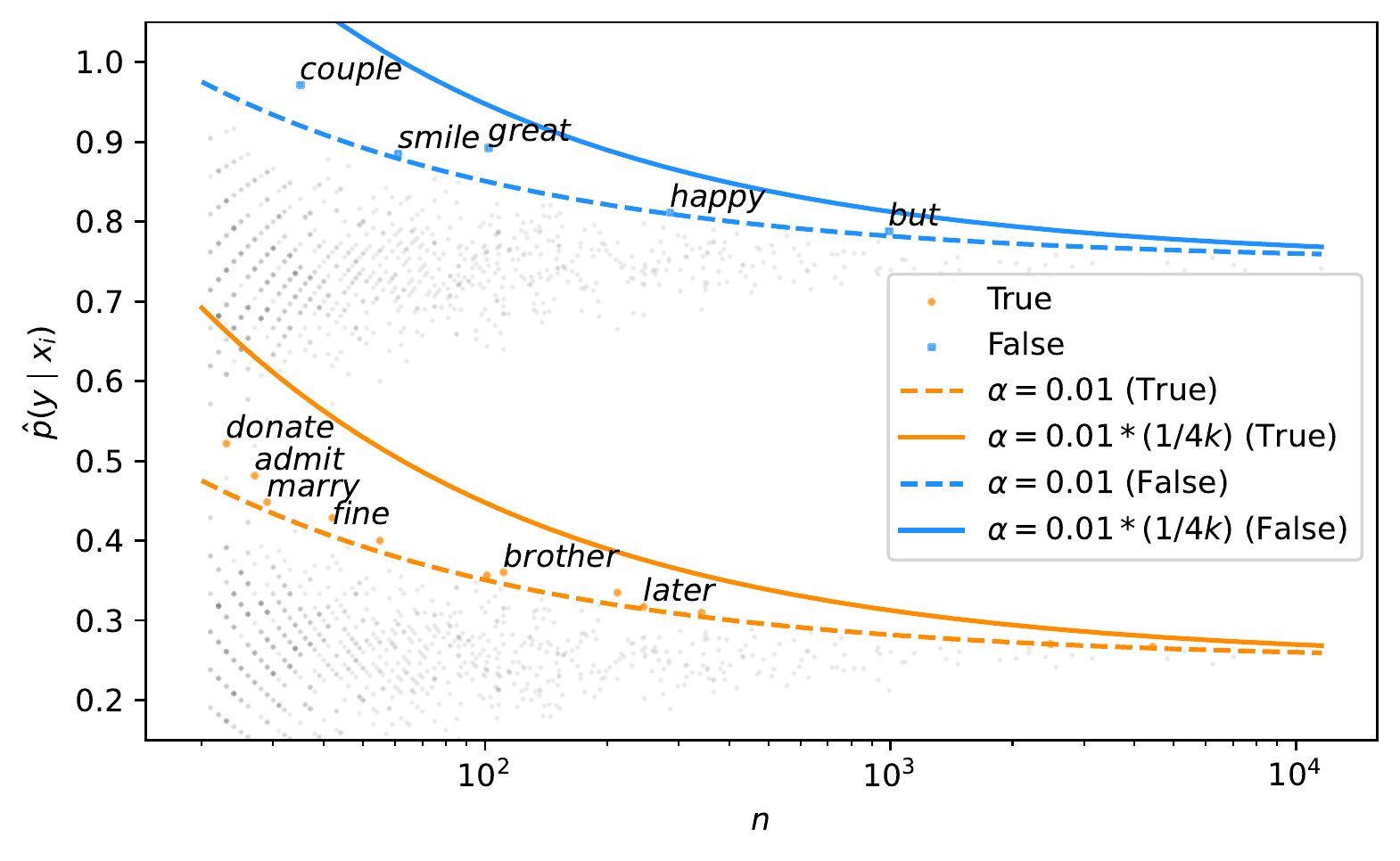}
        \caption{Possible Stories (ours)}
    \end{subfigure}
    \begin{subfigure}{.49\textwidth}
        \centering
        \includegraphics[width=\linewidth]{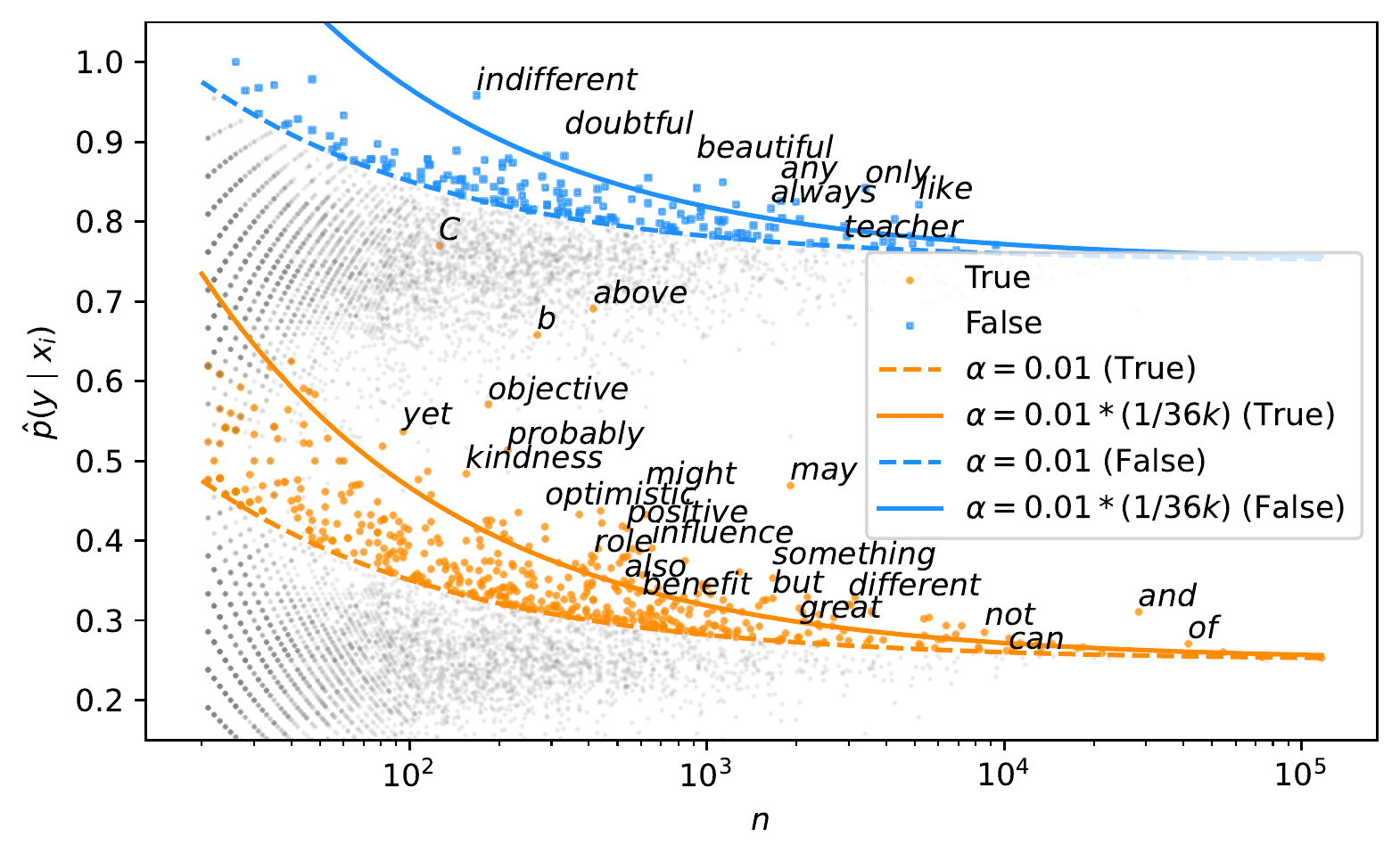}
        \caption{RACE}
    \end{subfigure} \\ \vspace{1em}
    \begin{subfigure}{.49\textwidth}
        \centering
        \includegraphics[width=\linewidth]{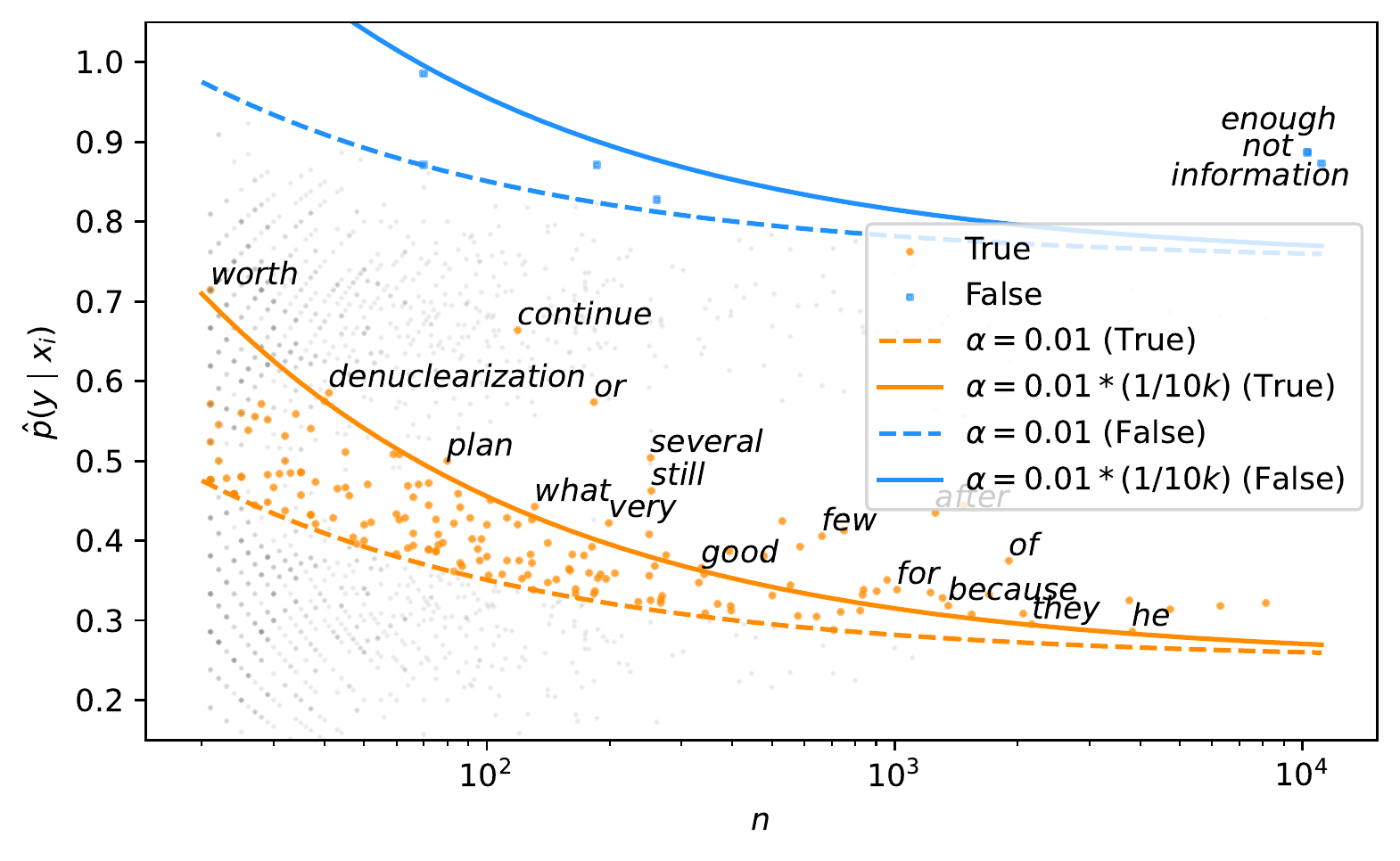}
        \caption{QuAIL}
    \end{subfigure}
    \begin{subfigure}{.49\textwidth}
        \centering
        \includegraphics[width=\linewidth]{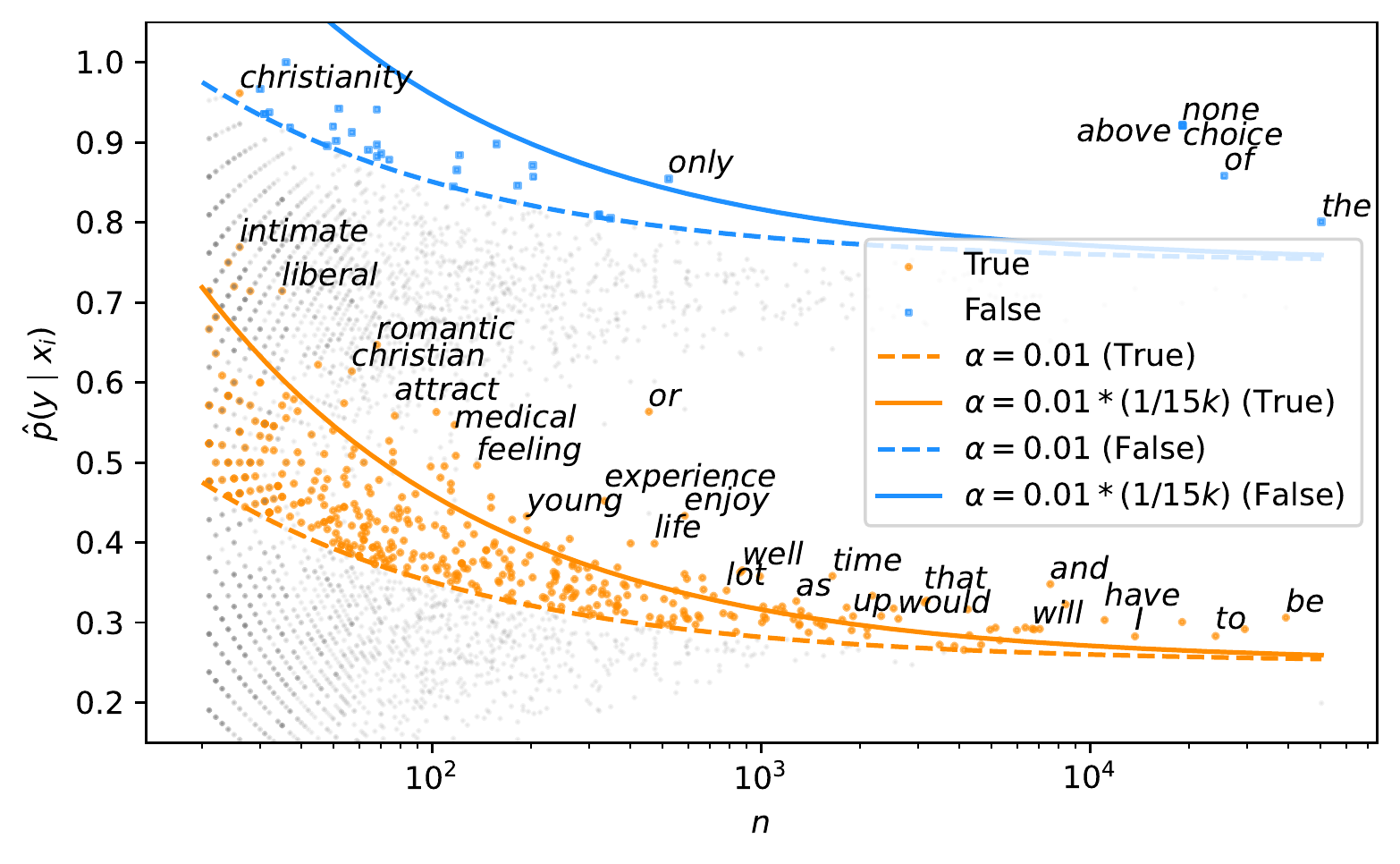}
        \caption{CosmosQA}
    \end{subfigure}
    \caption{
    Token-level annotation artifacts in the training examples of our dataset, RACE, QuAIL, and CosmosQA.
    All tokens are below $\alpha=0.01$ with a conservative Bonferroni correction for 3,990, 15,472, 35,762, and 9,688 vocabulary items, respectively.
    }
    \label{fig:artifact-four-datasets}
\end{figure*}

One of our motivations for crowdsourcing multiple questions for the same set of answer options is to minimize superficial patterns (i.e., annotation artifacts) in the collected examples, especially in their answer options. 
To validate this, we first compare the supervised performance in three ablation settings (no context passage, no question, and answer options only).
We use DeBERTa-large and report the test score on our dataset, RACE, QuAIL, and CosmosQA.
Table~\ref{tab:ablation-with-others} shows that although the no-context performance on our dataset is relatively high, the no-question and option-only performances are lower than the others.
This result implies that the question and answer options in our dataset are mutually indispensable for predicting the correct answer, while in the other datasets, the options on their own and their relationship with the context are informative for the prediction.

To visualize the actual tokens that create annotation artifacts, we follow \citet{gardner-etal-2021-competency}, who propose analyzing token-level features in terms of the empirical probability of labels $\hat{p}(y|x_i)$ given a specific token (vocabulary item) $x_i$ appearing in input $X$.
Here, the label $y$ indicates whether an answer option is the correct (True) or not (False).
We plot the probability $\hat{p}(y|x_i)$ and the number of occurrences ($n$) for the tokens of the training questions in our dataset, RACE, QuAIL, and CosmosQA (Figures~\ref{fig:artifact-four-datasets}) for comparison.
To see if the null-hypothesis (i.e., the token does not co-occur with a specific label) is accepted or rejected, we compute $z$-statistics and plot the level of statistical significance $\alpha=0.01$ and its conservative Bonferroni correction \cite{bonferroni1936teoria} for the vocabulary items ($\alpha=0.01/|V|$).
We find that for the true label in our dataset, only 12 tokens are above $\alpha=0.01$ and no tokens are above $\alpha=0.01/|V|$ where several content words, such as \textit{admit}, \textit{fine}, and \textit{great} are possibly helpful for predicting the correct label.
By contrast, 421 and 84 tokens in CosmosQA are found to be statistically significant at the respective levels, where many content words, such as \textit{enjoy} and \textit{life}, function words, such as \textit{or}, and the task-specific phrase (\textit{none of the above choice}) are strong indicators.
We observe similar trends for RACE and QuAIL. 
In Appendix~\ref{app:annotation-artifact}, we report the numbers of vocabulary items above the levels of statistical significance for the four datasets.
These results show that the answer options in our dataset contain minimal annotation artifacts compared to those of the other datasets.

\subsection{Question and Reasoning Types}
\label{sec:question_type}

\begin{figure}[t]
    \centering
    \includegraphics[width=0.9\linewidth]{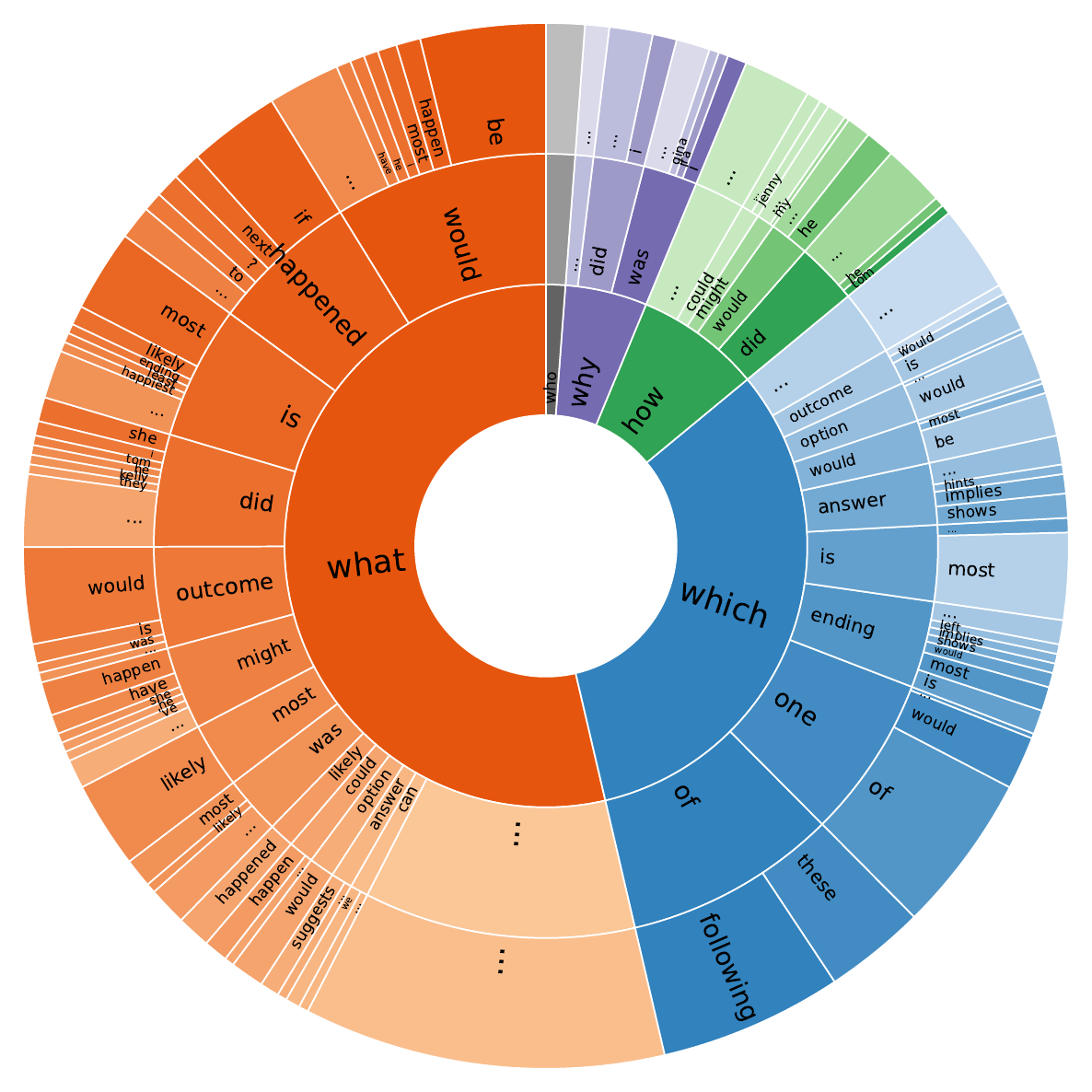}
    \caption{Question words and the subsequent words of the questions in our dataset.}
    \label{fig:qword-pie}
\end{figure}

The question words and subsequent words in the test questions are plotted in Figure~\ref{fig:qword-pie}.
We find that more than half the number of questions are \textit{what} questions, seemingly asking about the concrete content of the story.
We also observe subsequent words, such as \textit{would}, \textit{outcome}, and \textit{if}, which lead to a statement requiring commonsense reasoning.

\begin{figure}[t]
    \includegraphics[width=\linewidth]{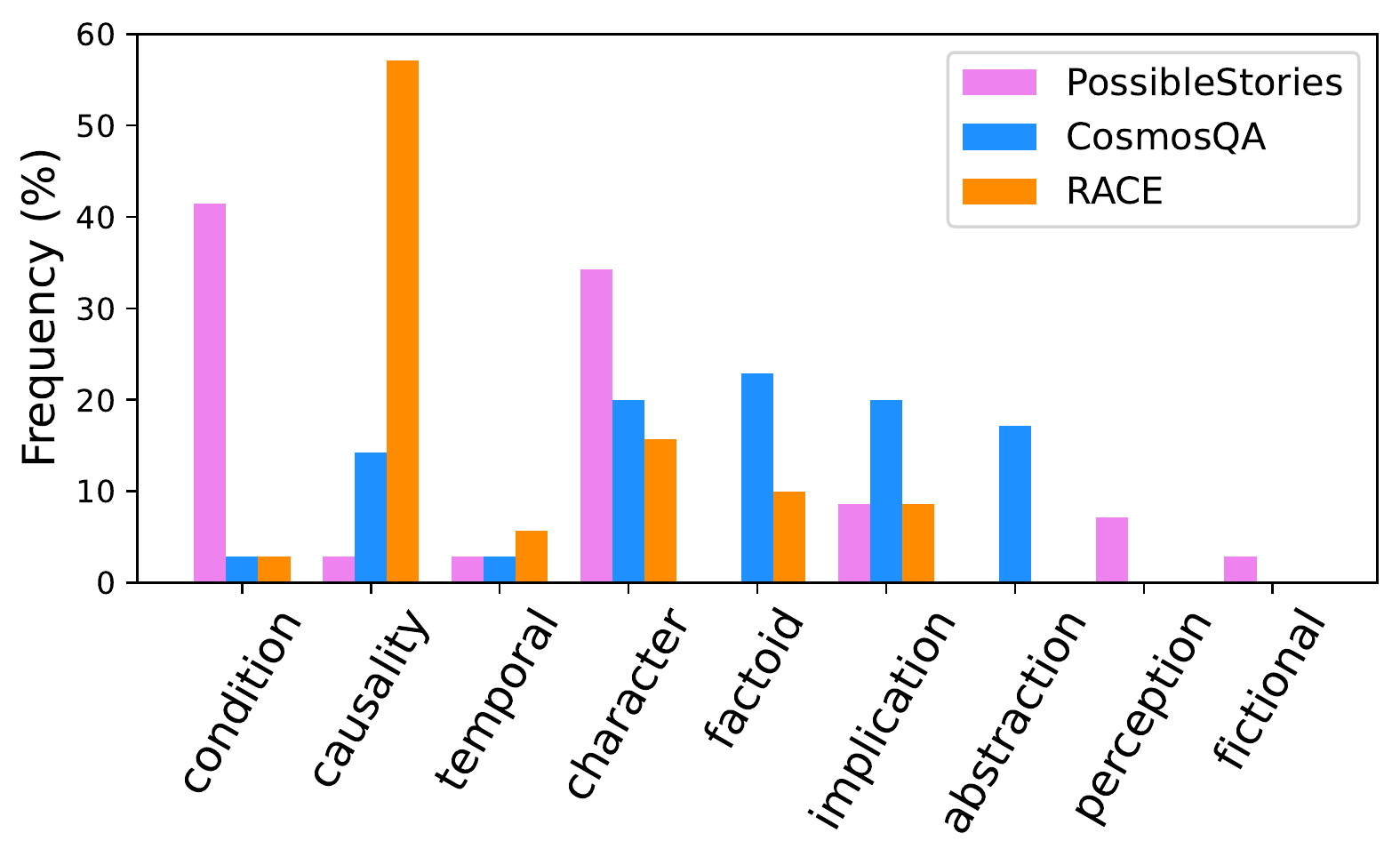}
    \caption{Reasoning types across our dataset, CosmosQA, and RACE.}
    \label{fig:reasoning-type-across-datasets}
\end{figure}

\begin{figure}[t]
    \includegraphics[width=\linewidth]{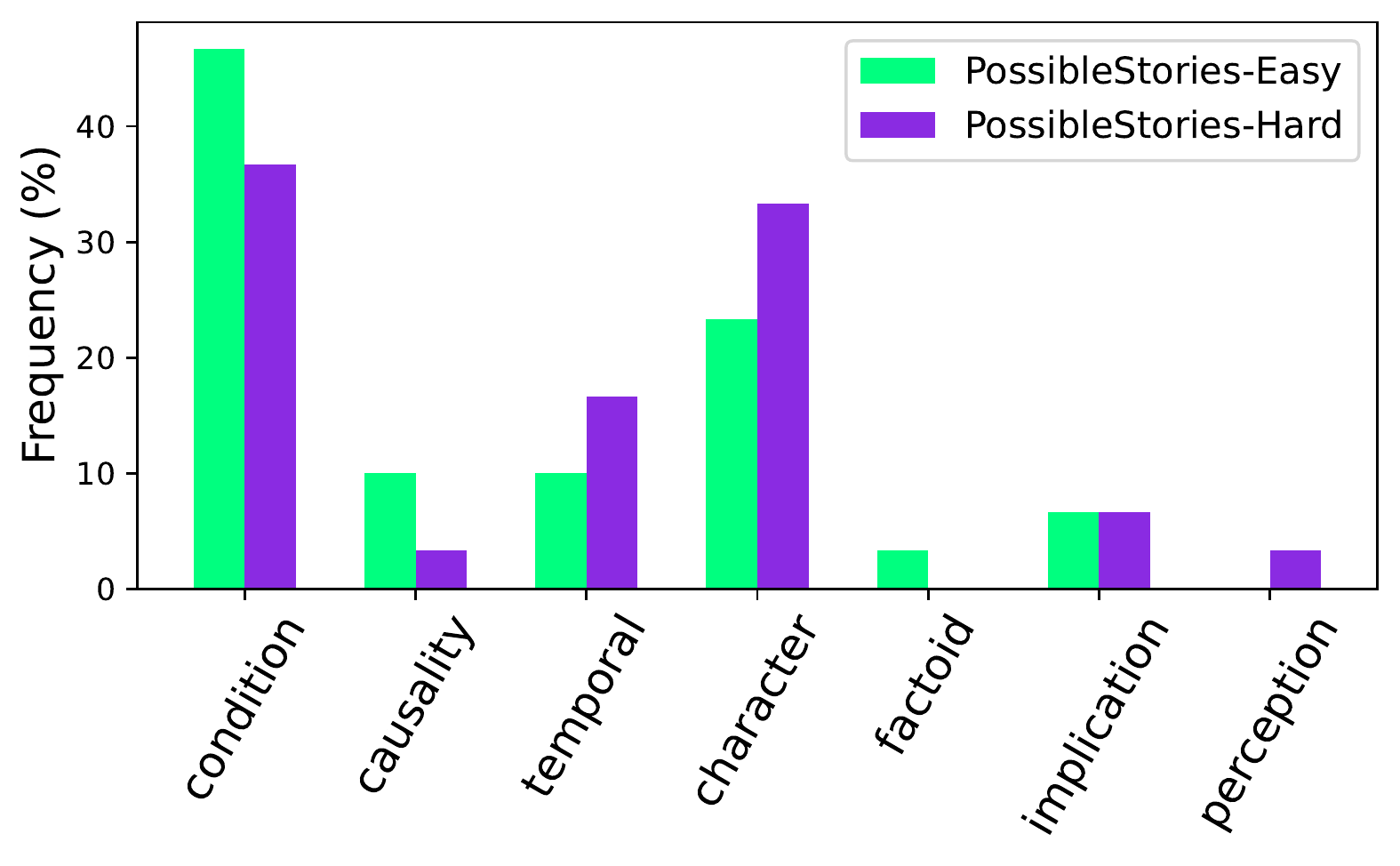}
    \caption{Reasoning types of easy and hard questions in our dataset.}
    \label{fig:reasoning-type-across-difficulty}
\end{figure}

\begin{figure*}[t]
    \setlength{\tabcolsep}{3pt}
    \small
    \begin{minipage}{\linewidth}\centering
    \fbox{\parbox{0.98\linewidth}{
        \textbf{P:} The Smith family loved to go on day trips on their boat in the summer. One day, they decided it would be fun to take the kids to a new place. They chose to travel north to a beach that wasn't terribly far away. The children had a wonderful time and met a new friend to play with.
        }%
    } \vspace{0.5em}
    \end{minipage}
    \begin{minipage}{\linewidth}
    \hphantom{Q}\textbf{Q1:} Which of these is the most negative ending? 
    \hphantom{Q}\textbf{Q2:} Which of these implies that the trip they took was successful? \\
    \hphantom{Q}\textbf{Q3:} Which ending implies the Smith kids were bad at staying in touch? 
    \hphantom{Q}\textbf{Q4:} Which ending involves the most conflict?
    \vspace{0.5em}
    \end{minipage}
    \def\arraystretch{1.15}
    \begin{tabular}{ccccp{0.8\linewidth}}
    Q1 & Q2 & Q3 & Q4 & Options \\
    \checkbox & \crossedbox & \checkbox & \checkbox & \textbf{A:} They kept in touch with their friend even after they went home. \\
    \checkedbox & \checkbox & \checkbox  & \checkedbox &\textbf{B:} At the end of the day the kids got into a fight with each other and were happy to leave. \\
    \checkbox & \checkedbox & \checkbox & \checkbox &\textbf{C:} The Smith's decided they'd visit a new beach every year, and they made tons of new friends. \\ 
    \crossedbox & \checkbox & \checkedbox & \crossedbox & \textbf{D:} They went home though and the kids never saw their friend again.
    \end{tabular}
    \caption{
    Example of questions with a single passage.
    Check mark (\colorcheckmark) indicates the correct option.
    Cross mark (\colorcrossmark) indicates that DeBERTa-large (v3) makes an incorrect prediction with that option.
    }
    \label{fig:error-example}
\end{figure*}

To investigate the kind of reasoning required for answering, we annotate the collected questions with reasoning types.
Considering previous studies, we define nine reasoning types (See Appendices~\ref{app:reasoning-type-definition} and \ref{app:reasoning-type-example} for the definitions and examples).
We annotate 70 questions from our test set and the same number of questions taken from CosmosQA and RACE for comparison (Figure~\ref{fig:reasoning-type-across-datasets}).
In addition, to examine the relationship between question difficulty and reasoning types, we split the test examples into easy and hard subsets and annotate 30 questions for each subset (Figure~\ref{fig:reasoning-type-across-difficulty}).
The easy questions are those for which the human--model accuracy gap is 0\% in terms of accuracy, and the hard ones are those for which the gap is larger than 60\% (215 and 64 examples).
To compute model performance, we average the accuracy of the five models (BERT-base and large, RoBERTa-base and large, and RoBERTa-large, which is fine-tuned on RACE; all models are fine-tuned on our training set).
Apart from the reasoning types, we independently check whether each question requires counterfactual reasoning.
This includes not only the condition and fiction types but also other types such as temporal reasoning in the sense that it can be involved in reasoning over counterfactual conditions.
The frequency (\%) of such questions is as follows: ours 68.6, ours-hard 76.7, ours-easy 66.7, CosmosQA 4.3, and RACE 2.9.

In summary, our major findings are as follows:
\begin{itemize}[leftmargin=1em]
    \item Our dataset includes more questions regarding conditions and characters' motivations and reactions than the other datasets. It also has a small number of fictional and perception questions, while the others do not.
    \item We do not observe abstraction and factoid questions in the annotated examples. However, we find several abstraction questions in our test set, one of which is presented in Appendix~\ref{app:reasoning-type-example}.
    \item Questions regarding characters' motivations and reactions are relatively harder, while questions regarding causality, which do not require counterfactual reasoning in most cases in our annotation, are easier. This corresponds with the fact that we find more counterfactual questions in the hard questions than in the easy questions.
\end{itemize}

\subsection{Case Study}

We present examples in which the strongest model (DeBERTa-large) makes incorrect predictions (Figure~\ref{fig:error-example}).
A single worker writes Q1 and Q2 and another worker writes Q3 and Q4.
Q1 and Q4 are annotated as perceptions (\textit{the most negative ending} and \textit{the most conflict}).
It seems that the model struggles to compare which option is more negative between options B (\textit{got into a fight...were happy}) and D (\textit{never saw their friend}).
Q2 and Q3 are annotated as implications (\textit{ imply...}).
For Q2, we must infer that option C (e.g., \textit{made tons of new friends}) implies success, but option B (\textit{kept in touch with their friend}) might sound more successful to the model.
More examples of other reasoning types are provided in Appendix~\ref{app:reasoning-type-example}.

\section{Discussion}


Circumscribing commonsense reasoning from simple heuristics has been a long-standing problem in the field of artificial intelligence \cite{levesque2014behavior}.
Although the answer options in our dataset are free from annotation artifacts, our ablation analysis in Section~\ref{sec:experiments} also shows that the questions and answer options may still involve some clues that the models can exploit.
Further research is needed to explain how commonsense reasoning is distinguished from a set of simple heuristics in machines' situational understanding.

In the question-writing task, one of the workers addresses the difficulty of creating questions that cannot be answered without reading the passages and that it is even practically impossible unless asking the questions to small children.
This illustrates that humans may also use a small amount of information available to draw inferences.
This issue arises possibly because of our task formulation, answering which option is more plausible than the others.
It can be argued that modifying this task to a generative task \cite{chen-etal-2020-mocha} is one way to directly assess machines' commonsense reasoning ability, but it should be noted that this would entail some difficulties in the evaluation of generated answers.

In addition, exploring what kind of conditions narrow down the possibilities of consequences is important to effectively evaluate machines' situational understanding. 
Although several studies have captured the dynamics of conditions in moral and immoral settings~\citep{emelin-etal-2021-moral}, many other factors come into play in decision-making in our daily lives, such as feelings, personal beliefs, expectations from others, or even unconscious biases.

\section{Conclusion}

This paper proposes a new dataset, Possible Stories, consisting of 4.5K crowdsourced questions with 1.3K story passages to investigate whether machines can infer the most plausible ending among four possible endings under certain situations postulated by questions.
We discover that current strong pretrained language models struggle to answer questions consistently, showing a large accuracy gap compared with humans.
A comparison with existing multiple-choice datasets demonstrates that our questions contain minimal annotation artifacts in the answer options and require counterfactual reasoning as well as an understanding of characters' motivations and emotions, suggesting that our dataset can serve as a challenging benchmark for future commonsense reasoning studies.

\section*{Ethical Considerations}
This study aims to facilitate the scientific study of machines' situated commonsense reasoning.
We use crowdsourcing for our data collection, taking care to avoid the exploitation of workers and pay well above the U.S. federal minimum wage.
The details of worker recruitment and the payment process are described in Appendix~\ref{app:worker}.
We also validate that the examples in our dataset do not contain unfair or harmful content.
In this section, we report our observations regarding the validation task.
This study is approved by our internal review board.

\paragraph{Content Validation for Fair Representation}
During content validation (Section~\ref{sec:data-validation}), we find that the level of content to be flagged is not trivial. 
There is a question containing the phrase \textit{mainstream COVID-related propaganda}, and one of the workers told us that the worker was unsure if it should be flagged.
Another case involves a story ending that describes the cooking skills of a male character in a bad light.
Does this representation perpetuate the negative stereotype that men are bad at cooking?
To investigate this, we should dive deeper into the semantic plausibility learned in language models \citep{porada-etal-2021-modeling, pedinotti-etal-2021-cat}.
Unless the focus is on the domain of natural science, there is less agreement on what would lean in spreading desirable and undesirable content, and the borderline can change across time and place.
It should also be noted that the degree of sensitivity towards underspecified biases depends on individuals' imagination and empathy.
Future work can examine how to effectively moderate the dataset for fair and unbiased representation.

\paragraph{Limitations} 
One of the limitations of the study is ``limited diversity.'' 
We observe that some systemic biases during data collection.
One example concerns a story in which the protagonist missed breakfast on a day of work.
Many crowdworkers come up with the possibility of the girlfriend bringing the lunch to the protagonist's workplace (referred to as \textit{I} throughout the context), but no one assumes that the boyfriend will do the same.
These types of unconscious biases can accumulate in datasets.
In addition, our dataset is limited to English.

\section*{Acknowledgments}

The authors would like to thank the anonymous reviewers for their insightful comments.
This work was supported by JST PRESTO
Grant Number JPMJPR20C4 and JSPS KAKENHI Grant Number 22K17954.

\bibliography{anthology, custom}

\begin{thebibliography}{51}
\expandafter\ifx\csname natexlab\endcsname\relax\def\natexlab#1{#1}\fi

\bibitem[{Aggarwal et~al.(2021)Aggarwal, Mandowara, Agrawal, Khandelwal,
  Singla, and Garg}]{aggarwal-etal-2021-explanations}
Shourya Aggarwal, Divyanshu Mandowara, Vishwajeet Agrawal, Dinesh Khandelwal,
  Parag Singla, and Dinesh Garg. 2021.
\newblock \href {https://doi.org/10.18653/v1/2021.acl-long.238} {{E}xplanations
  for {C}ommonsense{QA}: {N}ew {D}ataset and {M}odels}.
\newblock In \emph{Proceedings of the 59th Annual Meeting of the Association
  for Computational Linguistics and the 11th International Joint Conference on
  Natural Language Processing (Volume 1: Long Papers)}, pages 3050--3065,
  Online. Association for Computational Linguistics.

\bibitem[{Bartolo et~al.(2020)Bartolo, Roberts, Welbl, Riedel, and
  Stenetorp}]{bartolo-etal-2020-beat}
Max Bartolo, Alastair Roberts, Johannes Welbl, Sebastian Riedel, and Pontus
  Stenetorp. 2020.
\newblock \href {https://doi.org/10.1162/tacl_a_00338} {Beat the {AI}:
  Investigating adversarial human annotation for reading comprehension}.
\newblock \emph{Transactions of the Association for Computational Linguistics},
  8:662--678.

\bibitem[{Bhagavatula et~al.(2020)Bhagavatula, Bras, Malaviya, Sakaguchi,
  Holtzman, Rashkin, Downey, tau Yih, and Choi}]{Bhagavatula2020Abductive}
Chandra Bhagavatula, Ronan~Le Bras, Chaitanya Malaviya, Keisuke Sakaguchi, Ari
  Holtzman, Hannah Rashkin, Doug Downey, Wen tau Yih, and Yejin Choi. 2020.
\newblock \href {https://openreview.net/forum?id=Byg1v1HKDB} {Abductive
  commonsense reasoning}.
\newblock In \emph{International Conference on Learning Representations}.

\bibitem[{Black et~al.(2021)Black, Gao, Wang, Leahy, and Biderman}]{gpt-neo}
Sid Black, Leo Gao, Phil Wang, Connor Leahy, and Stella Biderman. 2021.
\newblock \href {https://doi.org/10.5281/zenodo.5297715} {{GPT-Neo}: Large
  scale autoregressive language modeling with mesh-tensorflow}.

\bibitem[{Bonferroni(1936)}]{bonferroni1936teoria}
Carlo~E. Bonferroni. 1936.
\newblock \href {https://cir.nii.ac.jp/crid/1570009749360424576} {Teoria
  statistica delle classi e calcolo delle probabilita}.
\newblock \emph{Pubblicazioni del R. Istituto Superiore di Scienze Economiche e
  Commerciali di Firenze}, 8:3--62.

\bibitem[{Bowman and Dahl(2021)}]{bowman-dahl-2021-will}
Samuel~R. Bowman and George Dahl. 2021.
\newblock \href {https://doi.org/10.18653/v1/2021.naacl-main.385} {What will it
  take to fix benchmarking in natural language understanding?}
\newblock In \emph{Proceedings of the 2021 Conference of the North American
  Chapter of the Association for Computational Linguistics: Human Language
  Technologies}, pages 4843--4855, Online. Association for Computational
  Linguistics.

\bibitem[{Chen et~al.(2020{\natexlab{a}})Chen, Stanovsky, Singh, and
  Gardner}]{chen-etal-2020-mocha}
Anthony Chen, Gabriel Stanovsky, Sameer Singh, and Matt Gardner.
  2020{\natexlab{a}}.
\newblock \href {https://doi.org/10.18653/v1/2020.emnlp-main.528} {{MOCHA}: A
  dataset for training and evaluating generative reading comprehension
  metrics}.
\newblock In \emph{Proceedings of the 2020 Conference on Empirical Methods in
  Natural Language Processing (EMNLP)}, pages 6521--6532, Online. Association
  for Computational Linguistics.

\bibitem[{Chen et~al.(2020{\natexlab{b}})Chen, Jiang, Poliak, Sakaguchi, and
  Van~Durme}]{chen-etal-2020-uncertain}
Tongfei Chen, Zhengping Jiang, Adam Poliak, Keisuke Sakaguchi, and Benjamin
  Van~Durme. 2020{\natexlab{b}}.
\newblock \href {https://doi.org/10.18653/v1/2020.acl-main.774} {Uncertain
  natural language inference}.
\newblock In \emph{Proceedings of the 58th Annual Meeting of the Association
  for Computational Linguistics}, pages 8772--8779, Online. Association for
  Computational Linguistics.

\bibitem[{Devlin et~al.(2019)Devlin, Chang, Lee, and
  Toutanova}]{devlin-etal-2019-bert}
Jacob Devlin, Ming-Wei Chang, Kenton Lee, and Kristina Toutanova. 2019.
\newblock \href {https://doi.org/10.18653/v1/N19-1423} {{BERT}: Pre-training of
  deep bidirectional transformers for language understanding}.
\newblock In \emph{Proceedings of the 2019 Conference of the North {A}merican
  Chapter of the Association for Computational Linguistics: Human Language
  Technologies, Volume 1 (Long and Short Papers)}, pages 4171--4186,
  Minneapolis, Minnesota. Association for Computational Linguistics.

\bibitem[{Emelin et~al.(2021)Emelin, Le~Bras, Hwang, Forbes, and
  Choi}]{emelin-etal-2021-moral}
Denis Emelin, Ronan Le~Bras, Jena~D. Hwang, Maxwell Forbes, and Yejin Choi.
  2021.
\newblock \href {https://doi.org/10.18653/v1/2021.emnlp-main.54} {Moral
  stories: Situated reasoning about norms, intents, actions, and their
  consequences}.
\newblock In \emph{Proceedings of the 2021 Conference on Empirical Methods in
  Natural Language Processing}, pages 698--718, Online and Punta Cana,
  Dominican Republic. Association for Computational Linguistics.

\bibitem[{Forbes et~al.(2020)Forbes, Hwang, Shwartz, Sap, and
  Choi}]{forbes-etal-2020-social}
Maxwell Forbes, Jena~D. Hwang, Vered Shwartz, Maarten Sap, and Yejin Choi.
  2020.
\newblock \href {https://doi.org/10.18653/v1/2020.emnlp-main.48} {Social
  chemistry 101: Learning to reason about social and moral norms}.
\newblock In \emph{Proceedings of the 2020 Conference on Empirical Methods in
  Natural Language Processing (EMNLP)}, pages 653--670, Online. Association for
  Computational Linguistics.

\bibitem[{Gardner et~al.(2020)Gardner, Artzi, Basmov, Berant, Bogin, Chen,
  Dasigi, Dua, Elazar, Gottumukkala, Gupta, Hajishirzi, Ilharco, Khashabi, Lin,
  Liu, Liu, Mulcaire, Ning, Singh, Smith, Subramanian, Tsarfaty, Wallace,
  Zhang, and Zhou}]{gardner-etal-2020-evaluating}
Matt Gardner, Yoav Artzi, Victoria Basmov, Jonathan Berant, Ben Bogin, Sihao
  Chen, Pradeep Dasigi, Dheeru Dua, Yanai Elazar, Ananth Gottumukkala, Nitish
  Gupta, Hannaneh Hajishirzi, Gabriel Ilharco, Daniel Khashabi, Kevin Lin,
  Jiangming Liu, Nelson~F. Liu, Phoebe Mulcaire, Qiang Ning, Sameer Singh,
  Noah~A. Smith, Sanjay Subramanian, Reut Tsarfaty, Eric Wallace, Ally Zhang,
  and Ben Zhou. 2020.
\newblock \href {https://doi.org/10.18653/v1/2020.findings-emnlp.117}
  {Evaluating models{'} local decision boundaries via contrast sets}.
\newblock In \emph{Findings of the Association for Computational Linguistics:
  EMNLP 2020}, pages 1307--1323, Online. Association for Computational
  Linguistics.

\bibitem[{Gardner et~al.(2021)Gardner, Merrill, Dodge, Peters, Ross, Singh, and
  Smith}]{gardner-etal-2021-competency}
Matt Gardner, William Merrill, Jesse Dodge, Matthew Peters, Alexis Ross, Sameer
  Singh, and Noah~A. Smith. 2021.
\newblock \href {https://doi.org/10.18653/v1/2021.emnlp-main.135} {Competency
  problems: On finding and removing artifacts in language data}.
\newblock In \emph{Proceedings of the 2021 Conference on Empirical Methods in
  Natural Language Processing}, pages 1801--1813, Online and Punta Cana,
  Dominican Republic. Association for Computational Linguistics.

\bibitem[{Gururangan et~al.(2018)Gururangan, Swayamdipta, Levy, Schwartz,
  Bowman, and Smith}]{gururangan-etal-2018-annotation}
Suchin Gururangan, Swabha Swayamdipta, Omer Levy, Roy Schwartz, Samuel Bowman,
  and Noah~A. Smith. 2018.
\newblock \href {https://doi.org/10.18653/v1/N18-2017} {Annotation artifacts in
  natural language inference data}.
\newblock In \emph{Proceedings of the 2018 Conference of the North {A}merican
  Chapter of the Association for Computational Linguistics: Human Language
  Technologies, Volume 2 (Short Papers)}, pages 107--112, New Orleans,
  Louisiana. Association for Computational Linguistics.

\bibitem[{He et~al.(2021)He, Gao, and Chen}]{he2021debertav3}
Pengcheng He, Jianfeng Gao, and Weizhu Chen. 2021.
\newblock \href {https://arxiv.org/abs/2111.09543} {{DeBERTaV3}: Improving
  {DeBERTa} using {ELECTRA}-style pre-training with gradient-disentangled
  embedding sharing}.
\newblock {a}rXiv preprint 2111.09543.

\bibitem[{Huang et~al.(2019)Huang, Le~Bras, Bhagavatula, and
  Choi}]{huang-etal-2019-cosmos}
Lifu Huang, Ronan Le~Bras, Chandra Bhagavatula, and Yejin Choi. 2019.
\newblock \href {https://doi.org/10.18653/v1/D19-1243} {Cosmos {QA}: Machine
  reading comprehension with contextual commonsense reasoning}.
\newblock In \emph{Proceedings of the 2019 Conference on Empirical Methods in
  Natural Language Processing and the 9th International Joint Conference on
  Natural Language Processing (EMNLP-IJCNLP)}, pages 2391--2401, Hong Kong,
  China. Association for Computational Linguistics.

\bibitem[{Jia and Liang(2017)}]{jia-liang-2017-adversarial}
Robin Jia and Percy Liang. 2017.
\newblock \href {https://doi.org/10.18653/v1/D17-1215} {Adversarial examples
  for evaluating reading comprehension systems}.
\newblock In \emph{Proceedings of the 2017 Conference on Empirical Methods in
  Natural Language Processing}, pages 2021--2031, Copenhagen, Denmark.
  Association for Computational Linguistics.

\bibitem[{Kaushik et~al.(2021)Kaushik, Kiela, Lipton, and
  Yih}]{kaushik-etal-2021-efficacy}
Divyansh Kaushik, Douwe Kiela, Zachary~C. Lipton, and Wen-tau Yih. 2021.
\newblock \href {https://doi.org/10.18653/v1/2021.acl-long.517} {On the
  efficacy of adversarial data collection for question answering: Results from
  a large-scale randomized study}.
\newblock In \emph{Proceedings of the 59th Annual Meeting of the Association
  for Computational Linguistics and the 11th International Joint Conference on
  Natural Language Processing (Volume 1: Long Papers)}, pages 6618--6633,
  Online. Association for Computational Linguistics.

\bibitem[{Khashabi et~al.(2020)Khashabi, Min, Khot, Sabharwal, Tafjord, Clark,
  and Hajishirzi}]{khashabi-etal-2020-unifiedqa}
Daniel Khashabi, Sewon Min, Tushar Khot, Ashish Sabharwal, Oyvind Tafjord,
  Peter Clark, and Hannaneh Hajishirzi. 2020.
\newblock \href {https://doi.org/10.18653/v1/2020.findings-emnlp.171}
  {{UNIFIEDQA}: Crossing format boundaries with a single {QA} system}.
\newblock In \emph{Findings of the Association for Computational Linguistics:
  EMNLP 2020}, pages 1896--1907, Online. Association for Computational
  Linguistics.

\bibitem[{Kummerfeld(2021)}]{kummerfeld-2021-quantifying}
Jonathan~K. Kummerfeld. 2021.
\newblock \href {https://doi.org/10.18653/v1/2021.acl-short.44} {Quantifying
  and avoiding unfair qualification labour in crowdsourcing}.
\newblock In \emph{Proceedings of the 59th Annual Meeting of the Association
  for Computational Linguistics and the 11th International Joint Conference on
  Natural Language Processing (Volume 2: Short Papers)}, pages 343--349,
  Online. Association for Computational Linguistics.

\bibitem[{Lai et~al.(2017)Lai, Xie, Liu, Yang, and Hovy}]{lai-etal-2017-race}
Guokun Lai, Qizhe Xie, Hanxiao Liu, Yiming Yang, and Eduard Hovy. 2017.
\newblock \href {https://doi.org/10.18653/v1/D17-1082} {{RACE}: Large-scale
  {R}e{A}ding comprehension dataset from examinations}.
\newblock In \emph{Proceedings of the 2017 Conference on Empirical Methods in
  Natural Language Processing}, pages 785--794, Copenhagen, Denmark.
  Association for Computational Linguistics.

\bibitem[{Levesque(2014)}]{levesque2014behavior}
Hector~J. Levesque. 2014.
\newblock \href {https://doi.org/https://doi.org/10.1016/j.artint.2014.03.007}
  {On our best behaviour}.
\newblock \emph{Artificial Intelligence}, 212:27--35.

\bibitem[{Li et~al.(2020)Li, Khashabi, Khot, Sabharwal, and
  Srikumar}]{li-etal-2020-unqovering}
Tao Li, Daniel Khashabi, Tushar Khot, Ashish Sabharwal, and Vivek Srikumar.
  2020.
\newblock \href {https://doi.org/10.18653/v1/2020.findings-emnlp.311}
  {{UNQOVER}ing stereotyping biases via underspecified questions}.
\newblock In \emph{Findings of the Association for Computational Linguistics:
  EMNLP 2020}, pages 3475--3489, Online. Association for Computational
  Linguistics.

\bibitem[{Liu et~al.(2019{\natexlab{a}})Liu, Schwartz, and
  Smith}]{liu-etal-2019-inoculation}
Nelson~F. Liu, Roy Schwartz, and Noah~A. Smith. 2019{\natexlab{a}}.
\newblock \href {https://doi.org/10.18653/v1/N19-1225} {Inoculation by
  fine-tuning: A method for analyzing challenge datasets}.
\newblock In \emph{Proceedings of the 2019 Conference of the North {A}merican
  Chapter of the Association for Computational Linguistics: Human Language
  Technologies, Volume 1 (Long and Short Papers)}, pages 2171--2179,
  Minneapolis, Minnesota. Association for Computational Linguistics.

\bibitem[{Liu et~al.(2019{\natexlab{b}})Liu, Ott, Goyal, Du, Joshi, Chen, Levy,
  Lewis, Zettlemoyer, and Stoyanov}]{Liu2019RoBERTaAR}
Yinhan Liu, Myle Ott, Naman Goyal, Jingfei Du, Mandar Joshi, Danqi Chen, Omer
  Levy, Mike Lewis, Luke Zettlemoyer, and Veselin Stoyanov. 2019{\natexlab{b}}.
\newblock \href {https://arxiv.org/abs/1907.11692} {{RoBERTa}: A robustly
  optimized {BERT} pretraining approach}.
\newblock {a}rXiv preprint 1907.11692.

\bibitem[{Lourie et~al.(2021)Lourie, Le~Bras, Bhagavatula, and
  Choi}]{Lourie2021}
Nicholas Lourie, Ronan Le~Bras, Chandra Bhagavatula, and Yejin Choi. 2021.
\newblock \href {https://ojs.aaai.org/index.php/AAAI/article/view/17590}
  {{UNICORN} on {RAINBOW}: A universal commonsense reasoning model on a new
  multitask benchmark}.
\newblock \emph{Proceedings of the AAAI Conference on Artificial Intelligence},
  35(15):13480--13488.

\bibitem[{Marvin and Linzen(2018)}]{marvin-linzen-2018-targeted}
Rebecca Marvin and Tal Linzen. 2018.
\newblock \href {https://doi.org/10.18653/v1/D18-1151} {Targeted syntactic
  evaluation of language models}.
\newblock In \emph{Proceedings of the 2018 Conference on Empirical Methods in
  Natural Language Processing}, pages 1192--1202, Brussels, Belgium.
  Association for Computational Linguistics.

\bibitem[{Meissner et~al.(2021)Meissner, Thumwanit, Sugawara, and
  Aizawa}]{meissner-etal-2021-embracing}
Johannes~Mario Meissner, Napat Thumwanit, Saku Sugawara, and Akiko Aizawa.
  2021.
\newblock \href {https://doi.org/10.18653/v1/2021.acl-short.109} {Embracing
  ambiguity: {S}hifting the training target of {NLI} models}.
\newblock In \emph{Proceedings of the 59th Annual Meeting of the Association
  for Computational Linguistics and the 11th International Joint Conference on
  Natural Language Processing (Volume 2: Short Papers)}, pages 862--869,
  Online. Association for Computational Linguistics.

\bibitem[{Mostafazadeh et~al.(2016)Mostafazadeh, Chambers, He, Parikh, Batra,
  Vanderwende, Kohli, and Allen}]{mostafazadeh-etal-2016-corpus}
Nasrin Mostafazadeh, Nathanael Chambers, Xiaodong He, Devi Parikh, Dhruv Batra,
  Lucy Vanderwende, Pushmeet Kohli, and James Allen. 2016.
\newblock \href {https://doi.org/10.18653/v1/N16-1098} {A corpus and cloze
  evaluation for deeper understanding of commonsense stories}.
\newblock In \emph{Proceedings of the 2016 Conference of the North {A}merican
  Chapter of the Association for Computational Linguistics: Human Language
  Technologies}, pages 839--849, San Diego, California. Association for
  Computational Linguistics.

\bibitem[{Mostafazadeh et~al.(2020)Mostafazadeh, Kalyanpur, Moon, Buchanan,
  Berkowitz, Biran, and Chu-Carroll}]{mostafazadeh-etal-2020-glucose}
Nasrin Mostafazadeh, Aditya Kalyanpur, Lori Moon, David Buchanan, Lauren
  Berkowitz, Or~Biran, and Jennifer Chu-Carroll. 2020.
\newblock \href {https://doi.org/10.18653/v1/2020.emnlp-main.370} {{GLUCOSE}:
  {G}enera{L}ized and {CO}ntextualized story explanations}.
\newblock In \emph{Proceedings of the 2020 Conference on Empirical Methods in
  Natural Language Processing (EMNLP)}, pages 4569--4586, Online. Association
  for Computational Linguistics.

\bibitem[{Nangia et~al.(2021)Nangia, Sugawara, Trivedi, Warstadt, Vania, and
  Bowman}]{nangia-etal-2021-ingredients}
Nikita Nangia, Saku Sugawara, Harsh Trivedi, Alex Warstadt, Clara Vania, and
  Samuel~R. Bowman. 2021.
\newblock \href {https://doi.org/10.18653/v1/2021.acl-long.98} {What
  ingredients make for an effective crowdsourcing protocol for difficult {NLU}
  data collection tasks?}
\newblock In \emph{Proceedings of the 59th Annual Meeting of the Association
  for Computational Linguistics and the 11th International Joint Conference on
  Natural Language Processing (Volume 1: Long Papers)}, pages 1221--1235,
  Online. Association for Computational Linguistics.

\bibitem[{Pavlick and Kwiatkowski(2019)}]{pavlick-kwiatkowski-2019-inherent}
Ellie Pavlick and Tom Kwiatkowski. 2019.
\newblock \href {https://doi.org/10.1162/tacl_a_00293} {Inherent disagreements
  in human textual inferences}.
\newblock \emph{Transactions of the Association for Computational Linguistics},
  7:677--694.

\bibitem[{Pedinotti et~al.(2021)Pedinotti, Rambelli, Chersoni, Santus, Lenci,
  and Blache}]{pedinotti-etal-2021-cat}
Paolo Pedinotti, Giulia Rambelli, Emmanuele Chersoni, Enrico Santus, Alessandro
  Lenci, and Philippe Blache. 2021.
\newblock \href {https://doi.org/10.18653/v1/2021.starsem-1.1} {Did the cat
  drink the coffee? challenging transformers with generalized event knowledge}.
\newblock In \emph{Proceedings of *SEM 2021: The Tenth Joint Conference on
  Lexical and Computational Semantics}, pages 1--11, Online. Association for
  Computational Linguistics.

\bibitem[{Porada et~al.(2021)Porada, Suleman, Trischler, and
  Cheung}]{porada-etal-2021-modeling}
Ian Porada, Kaheer Suleman, Adam Trischler, and Jackie Chi~Kit Cheung. 2021.
\newblock \href {https://doi.org/10.18653/v1/2021.naacl-main.138} {Modeling
  event plausibility with consistent conceptual abstraction}.
\newblock In \emph{Proceedings of the 2021 Conference of the North American
  Chapter of the Association for Computational Linguistics: Human Language
  Technologies}, pages 1732--1743, Online. Association for Computational
  Linguistics.

\bibitem[{Qin et~al.(2019)Qin, Bosselut, Holtzman, Bhagavatula, Clark, and
  Choi}]{qin-etal-2019-counterfactual}
Lianhui Qin, Antoine Bosselut, Ari Holtzman, Chandra Bhagavatula, Elizabeth
  Clark, and Yejin Choi. 2019.
\newblock \href {https://doi.org/10.18653/v1/D19-1509} {Counterfactual story
  reasoning and generation}.
\newblock In \emph{Proceedings of the 2019 Conference on Empirical Methods in
  Natural Language Processing and the 9th International Joint Conference on
  Natural Language Processing (EMNLP-IJCNLP)}, pages 5043--5053, Hong Kong,
  China. Association for Computational Linguistics.

\bibitem[{Radford et~al.(2019)Radford, Wu, Child, Luan, Amodei, Sutskever
  et~al.}]{radford2019language}
Alec Radford, Jeffrey Wu, Rewon Child, David Luan, Dario Amodei, Ilya
  Sutskever, et~al. 2019.
\newblock \href {https://openai.com/blog/better-language-models/} {Language
  models are unsupervised multitask learners}.
\newblock \emph{OpenAI blog}, 1(8):9.

\bibitem[{Rashkin et~al.(2018)Rashkin, Bosselut, Sap, Knight, and
  Choi}]{rashkin-etal-2018-modeling}
Hannah Rashkin, Antoine Bosselut, Maarten Sap, Kevin Knight, and Yejin Choi.
  2018.
\newblock \href {https://doi.org/10.18653/v1/P18-1213} {Modeling naive
  psychology of characters in simple commonsense stories}.
\newblock In \emph{Proceedings of the 56th Annual Meeting of the Association
  for Computational Linguistics (Volume 1: Long Papers)}, pages 2289--2299,
  Melbourne, Australia. Association for Computational Linguistics.

\bibitem[{Reimers and Gurevych(2019)}]{reimers-gurevych-2019-sentence}
Nils Reimers and Iryna Gurevych. 2019.
\newblock \href {https://doi.org/10.18653/v1/D19-1410} {Sentence-{BERT}:
  Sentence embeddings using {S}iamese {BERT}-networks}.
\newblock In \emph{Proceedings of the 2019 Conference on Empirical Methods in
  Natural Language Processing and the 9th International Joint Conference on
  Natural Language Processing (EMNLP-IJCNLP)}, pages 3982--3992, Hong Kong,
  China. Association for Computational Linguistics.

\bibitem[{Rogers et~al.(2020)Rogers, Kovaleva, Downey, and
  Rumshisky}]{Rogers2020GettingCT}
Anna Rogers, Olga Kovaleva, Matthew Downey, and Anna Rumshisky. 2020.
\newblock \href {https://doi.org/10.1609/aaai.v34i05.6398} {Getting closer to
  {AI} complete question answering: A set of prerequisite real tasks}.
\newblock \emph{Proceedings of the AAAI Conference on Artificial Intelligence},
  34(05):8722--8731.

\bibitem[{Saha et~al.(2021)Saha, Yadav, Bauer, and
  Bansal}]{saha-etal-2021-explagraphs}
Swarnadeep Saha, Prateek Yadav, Lisa Bauer, and Mohit Bansal. 2021.
\newblock \href {https://doi.org/10.18653/v1/2021.emnlp-main.609}
  {{E}xpla{G}raphs: An explanation graph generation task for structured
  commonsense reasoning}.
\newblock In \emph{Proceedings of the 2021 Conference on Empirical Methods in
  Natural Language Processing}, pages 7716--7740, Online and Punta Cana,
  Dominican Republic. Association for Computational Linguistics.

\bibitem[{Sakaguchi et~al.(2020)Sakaguchi, Le~Bras, Bhagavatula, and
  Choi}]{sakaguchi-etal-2020-winogrande}
Keisuke Sakaguchi, Ronan Le~Bras, Chandra Bhagavatula, and Yejin Choi. 2020.
\newblock \href {https://doi.org/10.1609/aaai.v34i05.6399} {Wino{G}rande: An
  adversarial {W}inograd schema challenge at scale}.
\newblock \emph{Proceedings of the AAAI Conference on Artificial Intelligence},
  34(05):8732--8740.

\bibitem[{Sap et~al.(2019)Sap, Rashkin, Chen, Le~Bras, and
  Choi}]{sap-etal-2019-social}
Maarten Sap, Hannah Rashkin, Derek Chen, Ronan Le~Bras, and Yejin Choi. 2019.
\newblock \href {https://doi.org/10.18653/v1/D19-1454} {Social {IQ}a:
  Commonsense reasoning about social interactions}.
\newblock In \emph{Proceedings of the 2019 Conference on Empirical Methods in
  Natural Language Processing and the 9th International Joint Conference on
  Natural Language Processing (EMNLP-IJCNLP)}, pages 4463--4473, Hong Kong,
  China. Association for Computational Linguistics.

\bibitem[{Sugawara et~al.(2022)Sugawara, Nangia, Warstadt, and
  Bowman}]{sugawara-etal-2022-makes}
Saku Sugawara, Nikita Nangia, Alex Warstadt, and Samuel Bowman. 2022.
\newblock \href {https://aclanthology.org/2022.acl-long.479} {What makes
  reading comprehension questions difficult?}
\newblock In \emph{Proceedings of the 60th Annual Meeting of the Association
  for Computational Linguistics (Volume 1: Long Papers)}, pages 6951--6971,
  Dublin, Ireland. Association for Computational Linguistics.

\bibitem[{Talmor et~al.(2019)Talmor, Herzig, Lourie, and
  Berant}]{talmor-etal-2019-commonsenseqa}
Alon Talmor, Jonathan Herzig, Nicholas Lourie, and Jonathan Berant. 2019.
\newblock \href {https://doi.org/10.18653/v1/N19-1421} {{C}ommonsense{QA}: A
  question answering challenge targeting commonsense knowledge}.
\newblock In \emph{Proceedings of the 2019 Conference of the North {A}merican
  Chapter of the Association for Computational Linguistics: Human Language
  Technologies, Volume 1 (Long and Short Papers)}, pages 4149--4158,
  Minneapolis, Minnesota. Association for Computational Linguistics.

\bibitem[{Tandon et~al.(2019)Tandon, Dalvi, Sakaguchi, Clark, and
  Bosselut}]{tandon-etal-2019-wiqa}
Niket Tandon, Bhavana Dalvi, Keisuke Sakaguchi, Peter Clark, and Antoine
  Bosselut. 2019.
\newblock \href {https://doi.org/10.18653/v1/D19-1629} {{WIQA}: A dataset for
  {``}what if...{''} reasoning over procedural text}.
\newblock In \emph{Proceedings of the 2019 Conference on Empirical Methods in
  Natural Language Processing and the 9th International Joint Conference on
  Natural Language Processing (EMNLP-IJCNLP)}, pages 6076--6085, Hong Kong,
  China. Association for Computational Linguistics.

\bibitem[{Warstadt et~al.(2020)Warstadt, Parrish, Liu, Mohananey, Peng, Wang,
  and Bowman}]{warstadt-etal-2020-blimp-benchmark}
Alex Warstadt, Alicia Parrish, Haokun Liu, Anhad Mohananey, Wei Peng, Sheng-Fu
  Wang, and Samuel~R. Bowman. 2020.
\newblock \href {https://doi.org/10.1162/tacl_a_00321} {{BL}i{MP}: The
  benchmark of linguistic minimal pairs for {E}nglish}.
\newblock \emph{Transactions of the Association for Computational Linguistics},
  8:377--392.

\bibitem[{Williams et~al.(2018)Williams, Nangia, and
  Bowman}]{williams-etal-2018-broad}
Adina Williams, Nikita Nangia, and Samuel Bowman. 2018.
\newblock \href {https://doi.org/10.18653/v1/N18-1101} {A broad-coverage
  challenge corpus for sentence understanding through inference}.
\newblock In \emph{Proceedings of the 2018 Conference of the North {A}merican
  Chapter of the Association for Computational Linguistics: Human Language
  Technologies, Volume 1 (Long Papers)}, pages 1112--1122, New Orleans,
  Louisiana. Association for Computational Linguistics.

\bibitem[{Wolf et~al.(2020)Wolf, Debut, Sanh, Chaumond, Delangue, Moi, Cistac,
  Rault, Louf, Funtowicz, Davison, Shleifer, von Platen, Ma, Jernite, Plu, Xu,
  Le~Scao, Gugger, Drame, Lhoest, and Rush}]{wolf-etal-2020-transformers}
Thomas Wolf, Lysandre Debut, Victor Sanh, Julien Chaumond, Clement Delangue,
  Anthony Moi, Pierric Cistac, Tim Rault, Remi Louf, Morgan Funtowicz, Joe
  Davison, Sam Shleifer, Patrick von Platen, Clara Ma, Yacine Jernite, Julien
  Plu, Canwen Xu, Teven Le~Scao, Sylvain Gugger, Mariama Drame, Quentin Lhoest,
  and Alexander Rush. 2020.
\newblock \href {https://doi.org/10.18653/v1/2020.emnlp-demos.6} {Transformers:
  State-of-the-art natural language processing}.
\newblock In \emph{Proceedings of the 2020 Conference on Empirical Methods in
  Natural Language Processing: System Demonstrations}, pages 38--45, Online.
  Association for Computational Linguistics.

\bibitem[{Zellers et~al.(2019)Zellers, Holtzman, Bisk, Farhadi, and
  Choi}]{zellers-etal-2019-hellaswag}
Rowan Zellers, Ari Holtzman, Yonatan Bisk, Ali Farhadi, and Yejin Choi. 2019.
\newblock \href {https://doi.org/10.18653/v1/P19-1472} {{H}ella{S}wag: Can a
  machine really finish your sentence?}
\newblock In \emph{Proceedings of the 57th Annual Meeting of the Association
  for Computational Linguistics}, pages 4791--4800, Florence, Italy.
  Association for Computational Linguistics.

\bibitem[{Zhang and Choi(2021)}]{zhang-choi-2021-situatedqa}
Michael Zhang and Eunsol Choi. 2021.
\newblock \href {https://doi.org/10.18653/v1/2021.emnlp-main.586}
  {{S}ituated{QA}: Incorporating extra-linguistic contexts into {QA}}.
\newblock In \emph{Proceedings of the 2021 Conference on Empirical Methods in
  Natural Language Processing}, pages 7371--7387, Online and Punta Cana,
  Dominican Republic. Association for Computational Linguistics.

\bibitem[{Zhang et~al.(2017)Zhang, Rudinger, Duh, and
  Van~Durme}]{zhang-etal-2017-ordinal}
Sheng Zhang, Rachel Rudinger, Kevin Duh, and Benjamin Van~Durme. 2017.
\newblock \href {https://doi.org/10.1162/tacl_a_00068} {Ordinal common-sense
  inference}.
\newblock \emph{Transactions of the Association for Computational Linguistics},
  5:379--395.

\end{thebibliography}
\bibliographystyle{acl_natbib}

\appendix
\section{Selecting Stories from ROCStories}\label{sec:story_selection}

We choose stories with more than 45 words and endings with more than 5 words to avoid stories that are too short or generic. 
We consult the annotation of the \textsc{GLUCOSE} dataset \citep{mostafazadeh-etal-2020-glucose} and select stories whose ending is annotated by workers with ratings higher than 1. 
Future work could investigate the effect of different causal relations on creating conditions, as well as possible endings. 

\section{Crowdworker Recruitment and Payment}\label{app:worker}

We recruit writers via Amazon Mechanical Turk (MTurk).
The number of workers who participated in the study is listed in Table~\ref{tab:worker_stats}.
Before initiating the data collection procedure, we first run a qualification task to identify workers who can participate in data collection. 
This task is a short version of a part of the main task and is open to \textit{any} crowdworker without \textit{any} qualifications such as HIT acceptance rate or number of HIT accepted, which are commonly used thresholds in the NLP community's data collection process via MTurk. 
We adapt this qualification following the recommendation of \citet{kummerfeld-2021-quantifying} to avoid the exploitation of crowdworkers. 
He demonstrates that imposing these prepared criteria is not fair because crowdworkers need to work on poorly paid tasks to achieve those qualifications in most cases. 

We pay \$1.0 USD for an ending writing task, \$1.5 for a question writing task, and \$1.0 for a validation task, estimating the completion time to be less than 5, 7.5, and 5 mins respectively.
This adds to more than \$12, which is well above the U.S. federal minimum wage. 
We do not calculate the wage according to the cost of living in each country where the workers reside, as we do not ask them where they live.

\begin{table}[t]
    \centering
    \begin{tabular}{p{3cm}rrr}
    \toprule
       &  Ending & Ques. & Valid. \\\midrule
       \# of crowdworkers & 163 & 66 & 65\\
       Ave. \# of examples & 54.6 & 79.9 & 243.3\\
       Max. \# of examples & 132 & 170 & 400 \\\bottomrule
    \end{tabular}
    \caption{Statistics of crowdworkers that participate in each task, the average and maximum number of generated examples per crowdworker.}
    \label{tab:worker_stats}
\end{table}

\section{Validation Results and Quality Control}
\label{app:validation}

\paragraph{Validation Results}
During the question-answer validation, 13.8\% of the collected questions are discarded.
Out of the four additional options, questions with no answer account for 1.8\%, those with more than two possible answers account for 6.8\%, ill-formed questions account for 1.8\%, and others account for 2.1\% of the total.
The high frequency of questions with more than two options is understood to be due to the possibility that some answer options are too similar to each other to create questions with a single correct answer.
Through the content validation process, 0.2\% of the questions are discarded.

\paragraph{Quality Control}
During the data collection process, we repeat all tasks three times (i.e., three batches).
The first and second batches have no workers in common, resulting in 52\% of the final dataset with a total of 66 workers.
For the final batch, we further qualify the workers who participated in these batches using three criteria: 1) writing more than nine questions, 2) mean human validation accuracy of more than 66\%, and 3) creating more than 90\% of questions as\textit{wh}-questions to ensure dataset quality.
Additionally, we manually check the comments given to each worker and exclude workers who tend to produce yes/no questions and those containing unethical or politically sensitive topics.
The final batch yields 48\% of the final dataset with 38 workers.

\section{Comparison of Models Fine-tuned on RACE and CosmosQA}
\label{app:cosmosqa-finetune}

\begin{table}[t]
 \centering
    \begin{tabular}{lcc} \toprule
    Model & RACE & CosmosQA \\ \midrule
    DeBERTa-large$^\dagger$ & 92.1 & 89.7 \\
    DeBERTa-large & 88.5 & 51.3 \\
    RoBERTa-large$^\dagger$ & 83.5 & 83.3 \\
    RoBERTa-large & 50.5 & 38.3 \\
    \bottomrule
    \end{tabular}
    \caption{
        Accuracy (\%) of models on our test set that are fine-tuned on RACE and CosmosQA respectively. $^\dagger$ indicates that the model is trained on our training set (i.e., supervised).
    }
    \label{tab:race-vs-cosmos}
\end{table}

In our experiments, we use RACE for fine-tuning our pretrained language models to adapt them to the multiple-choice task.
This is because we observe that RoBERTa-large and DeBERTa-large fine-tuned on RACE show higher performance than the corresponding models fine-tuned on CosmosQA (Table~\ref{tab:race-vs-cosmos}) in both unsupervised and supervised settings.

\section{Details of Experiments}
\label{app:experiments-detail}

\begin{table}[t]
    \centering
    \begin{tabular}{lcc}
    \toprule
    Model & $b$ & $lr$ \\ \midrule
    DeBERTa-large & 24 & 1e-5 \\
    DeBERTa-base & 48 & 3e-5 \\
    RoBERTa-large & 24 & 1e-5 \\
    RoBERTa-base & 48 & 3e-5 \\
    BERT-large & 36 & 1e-5 \\
    BERT-base & 72 & 3e-5 \\ \bottomrule
    \end{tabular}
    \caption{
        Hyperparameters used in the experiments. $b$ and $lr$ indicate the batch size and learning rate, respectively.
    }
    \label{tab:hyper}
\end{table}

Table~\ref{tab:hyper} reports the hyperparameters used in our experiments.
We use Huggingface's Transformers library \cite{wolf-etal-2020-transformers} for our experiments.

\begin{table*}[t]
    \centering
    \begin{tabular}{clrrrrrr} \toprule
    FT & Model & \multicolumn{2}{c}{Full input} &\multicolumn{2}{c}{No passage}& \multicolumn{2}{c}{No question}\\
 && Accuracy  & Consist. & Accuracy & Consist. & Accuracy & Consist. \\\midrule
    \crossmark & DeBERTa-large$^*$ & 60.2$\pm$1.7 & 19.9$\pm$2.2 & 58.1$\pm$2.6 & 19.9$\pm$1.7 & 21.8$\pm$1.6 & 0.5$\pm$0.4 \\
    \checkmark & DeBERTa-large$^*$ & 92.1$\pm$0.6 & 74.7$\pm$2.3 & 87.0$\pm$0.7 & 62.1$\pm$1.8 & 31.8$\pm$1.6 & 1.9$\pm$0.7 \\
    \bottomrule
    \end{tabular}
    \caption{
        Unsupervised and supervised performance (\%) with the standard deviations of DeBERTa-large in five runs.
        The five models are fine-tuned on RACE with different random seeds, respectively.
    }
    \label{tab:std}
\end{table*}

Table~\ref{tab:std} reports the detailed results of DeBERTa-large (fine-tuned on RACE) on our test set in the unsupervised and supervised settings.
Owing to computational constraints, we conduct five different runs only for this model, which is the strongest among the models we use in our experiments.
We do not observe large deviations across the runs.

\section{Annotation Artifacts in Answer Options}
\label{app:annotation-artifact}

\begin{table}[t] \setlength{\tabcolsep}{4pt}
    \centering
    \begin{tabular}{lcccc} \toprule
    $\alpha$ & Ours & Cosmos & RACE & QuAIL \\ \midrule
    $0.01$ & 12/5 & 421/33 & 475/163 & 173/7 \\
    $0.01/|V|$ & 0/0 & 84/6 & 104/19 & 39/3 \\ \midrule
    $|V|$ & 3,990 & 15,472 & 35,762 & 9,688 \\ \bottomrule
    \end{tabular}
    \caption{
        The number of vocabulary items that appear in correct/incorrect options above the levels of statistical significance ($\alpha=0.01$ and its conservative Bonferroni correction for the size of vocabulary $|V|$).
    }
    \label{tab:artifact-numbers}
\end{table}

We report the number of examples above different levels of statistical significance across the four analyzed datasets in Table~\ref{tab:artifact-numbers}.
The number for our dataset above $\alpha=0.01/|V|$ is zero, whereas those for the other datasets are significantly larger.
This result shows that our dataset does not suffer from token-level annotation artifacts in the answer options, supporting our findings on the option-only training results in Section~\ref{sec:performance-gap}.

\section{Definitions of Reasoning Types}
\label{app:reasoning-type-definition}

\begin{table}[t]
\centering \fontsize{10pt}{11.5pt}\selectfont
\setlength{\tabcolsep}{2.5pt}
\def\arraystretch{1.2}
\newcommand{\emphitem}[1]{\textbf{#1}}  
\begin{tabular}{lp{0.9\linewidth}} \toprule
    1. & \emphitem{Condition}: pre/post counterfactual conditions introduced in the question. \\
    2. & \emphitem{Causality}: causes and effects of events. \\
    3. & \emphitem{Temporal}: temporal relations between events. \\
    4. & \emphitem{Character}: characters' emotions, motivations, and reactions. \\
    5. & \emphitem{Factoid}: extracting entities from the context.\\
    6. & \emphitem{Abstraction}: lesson, conclusion, and summary of the context. \\
    7. & \emphitem{Implication}: paraphrasing and implication about events. \\
    8. & \emphitem{Perception}: reader's perceptual responses. \\
    9. & \emphitem{Fictional}: fictional situations as counterfactual condition. \\
    \bottomrule
\end{tabular}
\caption{Definitions of reasoning types.}
\label{tab:reasoning-type}
\end{table}

Annotating reasoning types is not a trivial task, particularly because the questions are fully written by humans without templates.
Moreover, it is possible to use many classification methods, and there is rarely a consensus on reasoning types.
For example, CosmosQA proposes seven reasoning types: pre-/post-conditions, motivations, reactions, temporal events, situational facts, counterfactuals, and other (e.g., cultural norms). 
In QuAIL, nine types of reasoning are proposed spanning three categories: temporal, factoid, character properties for \textit{text-based questions}, coreference, causality, belief states, subsequent entity states, event durations for \textit{questions that require world knowledge}, and \textit{unanswerable}.
After categorizing these into five types, we add three types: abstraction (summarizing what happened), implication (paraphrasing), and readers' (observers') perceptions. 
In addition, we differentiate reasoning over fiction from counterfactual in that it is a more specific type of counterfactual that is considered implausible for most people in the real world.
This results in the nine reasoning types listed in Table~\ref{tab:reasoning-type}.
Appendix~\ref{app:reasoning-type-example} presents some examples.

\section{More Examples for Reasoning Types and Difficulty}
\label{app:reasoning-type-example}

\begin{table}[t]
    \centering
    \begin{tabular}{p{0.2\linewidth}p{0.7\linewidth}}
    \toprule
    Reasoning &  Example \\\midrule
    Condition & Jeff is a child with a very vivid sense of imagination.  What is most likely to have happened next? \\
    Causality & Which is the most likely caused the guests to avoid shards of glass? \\ 
    Temporal & Which is most likely if Chris later felt sick to his stomach?\\ 
    Character & What outcome would be most upsetting to Ben?\\
    Factoid & Where did people hide the money they got? \\
    Abstraction & What lesson did she learn from the passage? \\
    Implication & Which answer implies Bob was pleased with his performance?\\
    Perception  & What is the most moral decision for Danielle? \\
    Fictional & How does Dylan get home? \\
     \bottomrule
    \end{tabular}
    \caption{
    Reasoning types we use in the annotation and their example questions.
    }
    \label{tab:question_type}
\end{table}

\begin{figure*}[t]
    \setlength{\tabcolsep}{3pt}
     \fontsize{10pt}{11.5pt}\selectfont
    \begin{minipage}{\linewidth}\centering
    \fbox{\parbox{0.98\linewidth}{ 
        \textbf{P1:} Lydia was listening to an old CD her boyfriend had burned for her. Her CD player was old but still working alright. She had lost track of her thoughts and was enjoying the music. Suddenly, the CD skipped out and stopped playing.
        }%
    } \vspace{0.5em}
    \end{minipage}
    \begin{minipage}{\linewidth}
    \textbf{Q1:} Why was the CD player unable to function? (causality, easy) \\
    \textbf{Q2:} Which answer indicates that Lydia would never be able to listen to the CD again? (implication, hard) \\
    \textbf{Q3:}~Which of the following is likely to occur if we know Lydia has realized the CD player cannot be fixed? \hphantom{Q3: }(condition, hard)
    \end{minipage}
    \def\arraystretch{1.15}
    \begin{tabular}{ccccp{0.8\linewidth}}
    Q1 & Q2 & Q3 & \hphantom{Q4} & Options \\
    \checkbox & \checkedbox & \checkbox & & \textbf{A:}  Lydia tried to fix it but the CD had a huge scratch. \\
    \checkbox & \checkbox & \checkbox  & &\textbf{B:} Lydia tinkered with the CD player and got it working again. \\
    \checkbox & \checkbox & \checkedbox & &\textbf{C:} Lydia went to bed upset, knowing she had to buy a new one in the morning. \\
    \checkedbox & \crossedbox & \crossedbox & & \textbf{D:} She realized the batteries in her CD player had died. \\
    \end{tabular}
    \vspace{0.5em}

    \begin{minipage}{\linewidth}\centering
    \fbox{\parbox{0.98\linewidth}{  
        \textbf{P2:} Darrel was waiting in the drive through for half an hour. He had about lost his patience. When he finally got to the window he was about to scream at them. They immediately apologized before he could.
        }%
    } \vspace{0.5em}
    \end{minipage}
    \begin{minipage}{\linewidth}
    \textbf{Q1:} How did the employees react when they saw Darrel's face turn red at the drive-through window? (character) \\
    \textbf{Q2:} How did Darrel respond after the employees apologized for the long wait? (character, easy) \\
    \textbf{Q3:}~If Darrel's mind was soon preoccupied with something entirely different, what was most likely to have \hphantom{Q3: }happened? (condition, easy) \\
    \textbf{Q4:} In this scenario, what most likely happened if Darrel was pleased soon thereafter? (character, hard)
    \end{minipage}
    
    \def\arraystretch{1.15}
    \begin{tabular}{ccccp{0.8\linewidth}}
    Q1 & Q2 & Q3 & Q4 & Options \\
    \checkbox & \checkbox & \checkbox & \checkedbox & \textbf{A:} They had an accident and offered free food to make it up to him. \\
    \checkbox & \checkedbox & \checkbox  & \checkbox &\textbf{B:} He chose not to accept the apology and asked to speak to the manager. \\
    \checkedbox & \checkbox & \checkbox & \crossedbox &\textbf{C:} They quickly gave him his food and informed him that there were very few employees \hphantom{C: }working that day. \\
    \checkbox & \checkbox & \checkedbox & \checkbox & \textbf{D:} Before he could open his mouth, his engine started smoking and he had to call a tow \hphantom{D: }truck. \\
    \end{tabular}
    \vspace{0.5em}
    
    \begin{minipage}{\linewidth}\centering
    \fbox{\parbox{0.98\linewidth}{ 
        \textbf{P3:} Jan checked to make sure no one was around. Her two older brothers had been sneaking around the garden lately. Being a curious child, Jan wanted to know what they were up to. She carefully opened the door to her brother's room.
        }%
    } \vspace{0.5em}
    \end{minipage}
    \begin{minipage}{\linewidth}
    \textbf{Q1:}~If Jan smelled pleasant aromas and felt fresh air in the room, what did she likely discover? (condition) \\
    \textbf{Q2:}~What was the likely outcome if Jan was left still feeling clueless about what her brothers had been up to? \hphantom{Q2: }(character) \\
    \textbf{Q3:}~Which outcome is the most unlikely to occur in reality? (fiction) \\
    \textbf{Q4:}~Which would be particularly unpleasant for Jan if she suffers from acute arachnophobia? (character)
    \end{minipage}
    \def\arraystretch{1.15}
    \begin{tabular}{ccccp{0.8\linewidth}}
    Q1 & Q2 & Q3 & Q4 & Options \\
    \checkbox & \checkbox & \checkbox & \checkedbox & \textbf{A:} Inside the back of their closet, she found several jars with spiders. \\
    \checkbox & \checkbox & \checkedbox  & \checkbox &\textbf{B:} There was a strange looking alien peeking out of a corner with fearful eyes. \\
    \checkedbox & \crossedbox & \checkbox & \checkbox &\textbf{C:} They had taken plants from the garden and moved them to their room. \\
    \checkbox & \checkedbox & \checkbox & \checkbox & \textbf{D:} The door slammed shut on her face as the cameras alerted the brothers of an intruder. \\
    \end{tabular}
    \vspace{0.5em}

    \begin{minipage}{\linewidth}\centering
    \fbox{\parbox{0.98\linewidth}{ 
        \textbf{P4:} Billy liked Christmas songs. But didn't know what a turtle dove was. He like turtle and knew they were green and had a shell. He also knew what a dove was, a type of bird.
        }%
    } \vspace{0.5em}
    \end{minipage}
    \begin{minipage}{\linewidth}
    \textbf{Q1:} What happened if it was the worst Christmas of Billy's life? (condition, easy) \\
    \textbf{Q2:} What happened if he pictured a turtle with wings? (fictional, easy) \\
    \textbf{Q3:} What outcome would be most tragic? (perception)
    \end{minipage}
    \def\arraystretch{1.15}
    \begin{tabular}{ccccp{0.8\linewidth}}
    Q1 & Q2 & Q3 & \hphantom{Q4} & Options \\
    \checkbox & \checkbox & \checkbox & & \textbf{A:} So he decided that 12 drummers drumming was a better part of the song. \\
    \checkbox & \checkedbox & \checkbox  & &\textbf{B:} He decided that a turtle dove was likely a flying turtle. \\
    \checkbox & \checkbox & \checkbox & &\textbf{C:} Billy became a famous author after embracing his love for holiday traditions. \\
    \checkedbox & \checkbox & \checkedbox & & \textbf{D:} He went to ask his mother about turtle doves, but when he found her in the bathtub, she was dead. \\
    \end{tabular}
    \caption{
    Examples in our dataset.
    Check mark (\colorcheckmark) indicates the correct option.
    Cross mark (\colorcrossmark) indicates that RoBERTa-large fine-tuned on RACE and our training set makes an incorrect prediction with that option.
    }
    \label{fig:other-examples}
\end{figure*}

The reasoning types and their example questions taken from our dataset are listed in Table~\ref{tab:question_type}.
We also show examples of passage, question, and answer options in our dataset, including easy and hard questions, in Figure~\ref{fig:other-examples}.
Each question ends with its reasoning type and easy/hard classification, if available.

\section{Annotation Instructions and Interfaces}
\label{app:annotation-instructions}

\subsection{Ending Writing}
\label{app:annotation-instructions-ending-writing}

\begin{figure*}[t]
    \includegraphics[width=\linewidth]{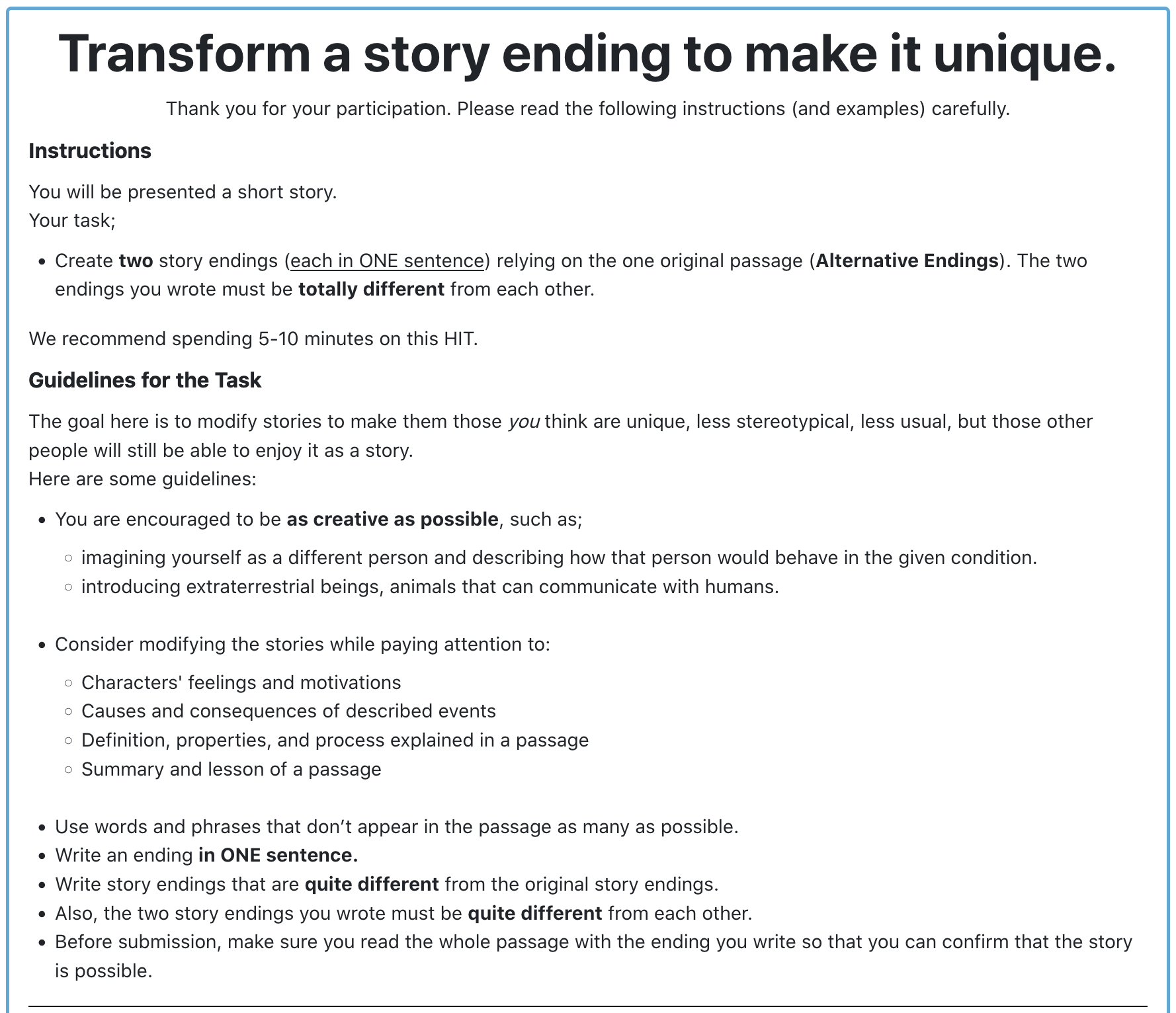}
    \caption{
        Instructions (1/2) used in the story ending writing task.
    }
    \label{fig:ending-writing-instructions1}
\end{figure*}

\begin{figure*}[t]
    \includegraphics[width=\linewidth]{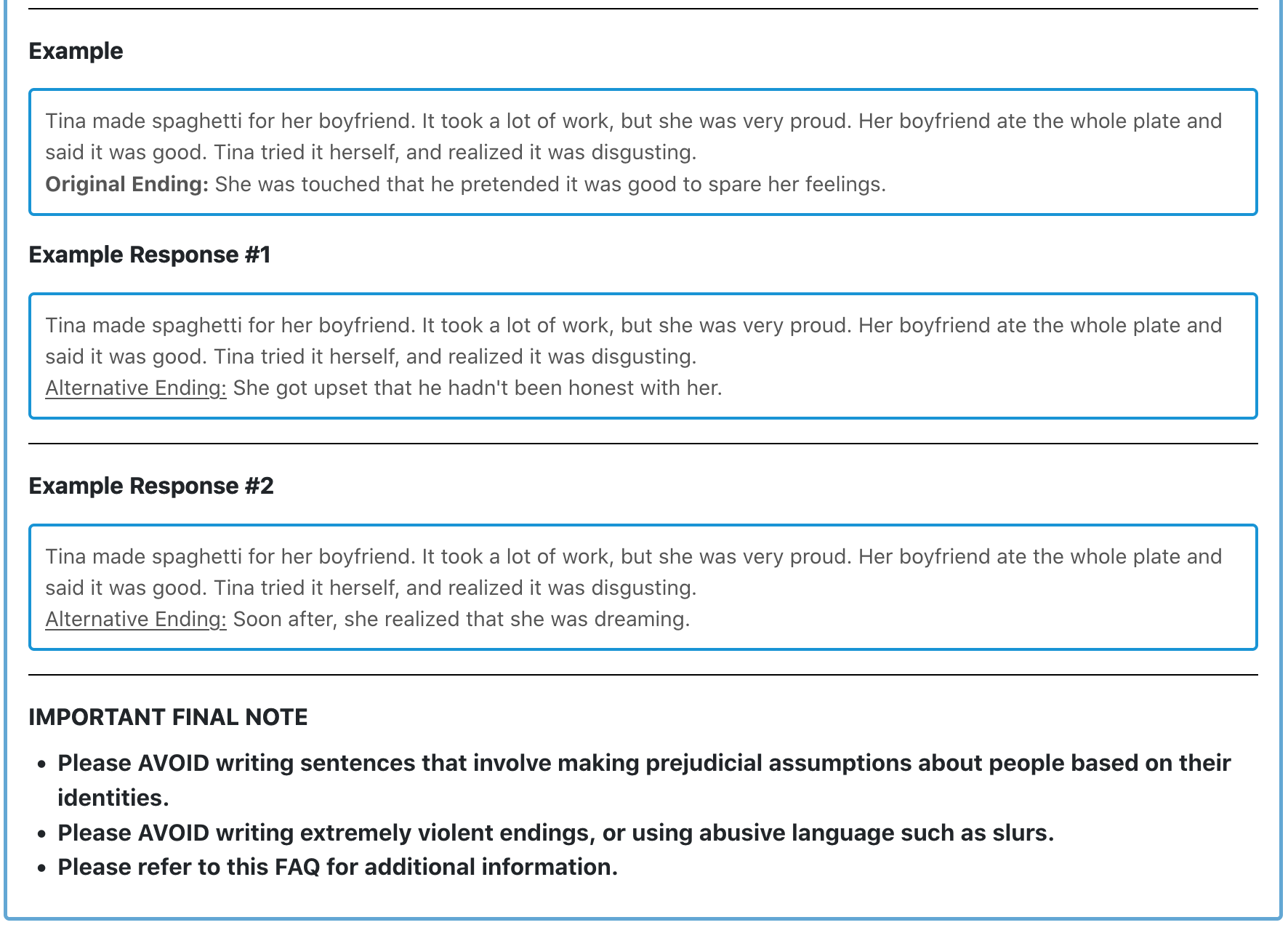}
    \caption{
        Instructions (2/2) used in the story ending writing task.
    }
    \label{fig:ending-writing-instructions2}
\end{figure*}

\begin{figure*}[t]
    \includegraphics[width=\linewidth]{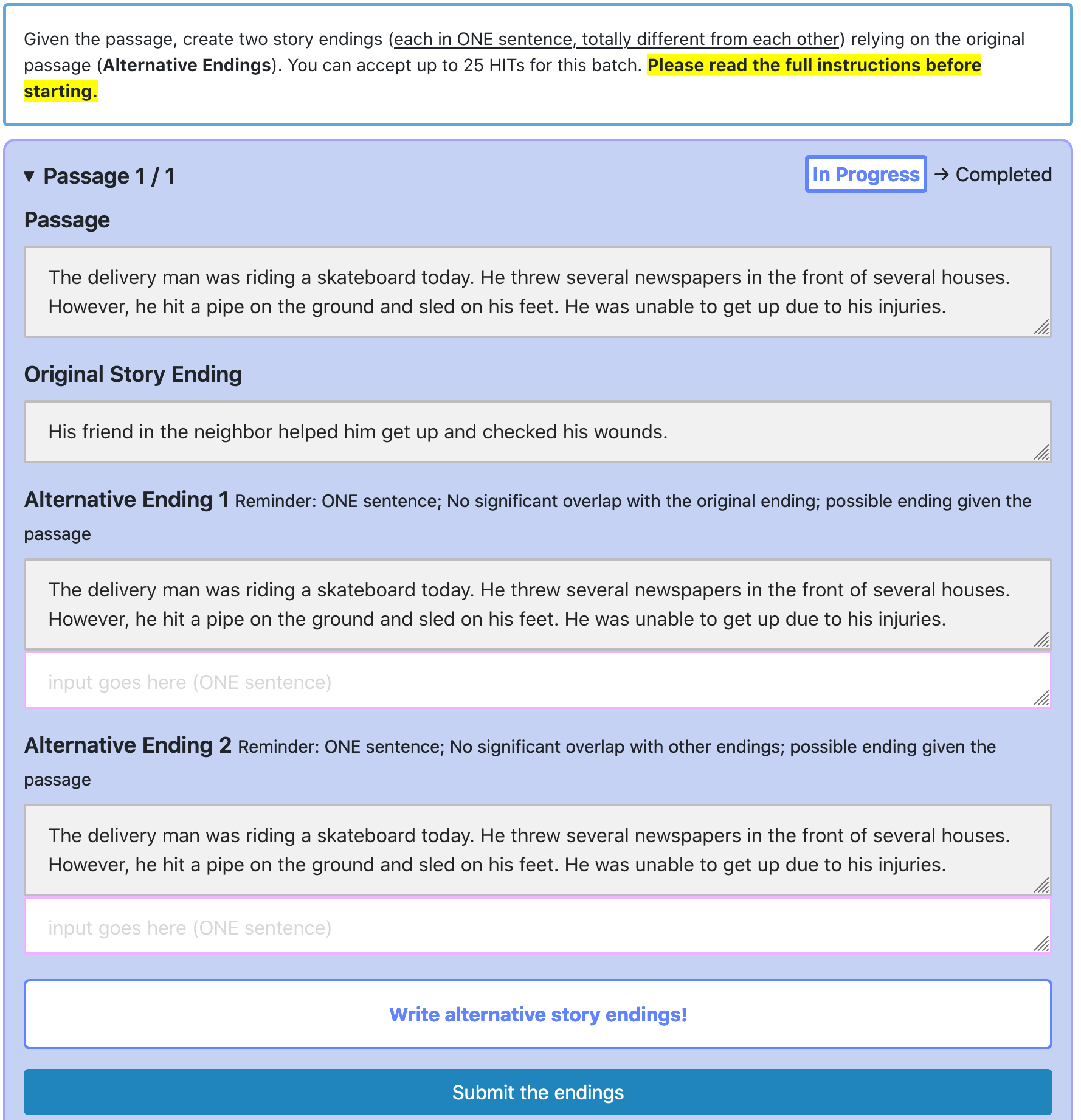}
    \caption{
        Interface used in the story ending writing task.
    }
    \label{fig:ending-writing-interface}
\end{figure*}

Figures~\ref{fig:ending-writing-instructions1} and \ref{fig:ending-writing-instructions2} show the instructions used in the story ending writing task.
Figure~\ref{fig:ending-writing-interface} shows the interface used in the story ending writing task.

\subsection{Question Writing}
\label{app:annotation-instructions-question-writing}

\begin{figure*}[t]
    \includegraphics[width=\linewidth]{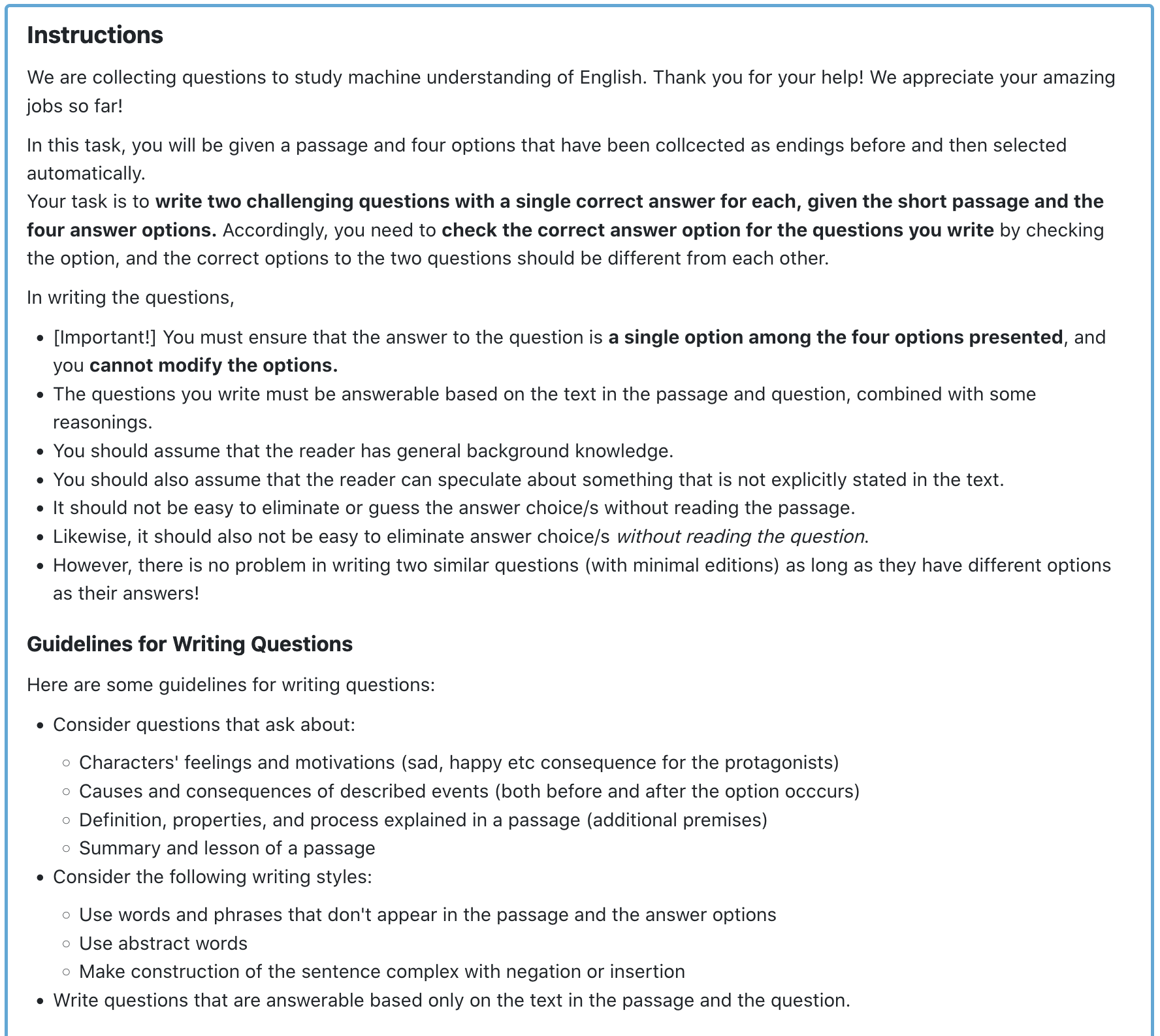}
    \caption{
        Instructions (1/3) used in the question writing task.
    }
    \label{fig:question-writing-instructions1}
\end{figure*}
\begin{figure*}[t]
    \includegraphics[width=\linewidth]{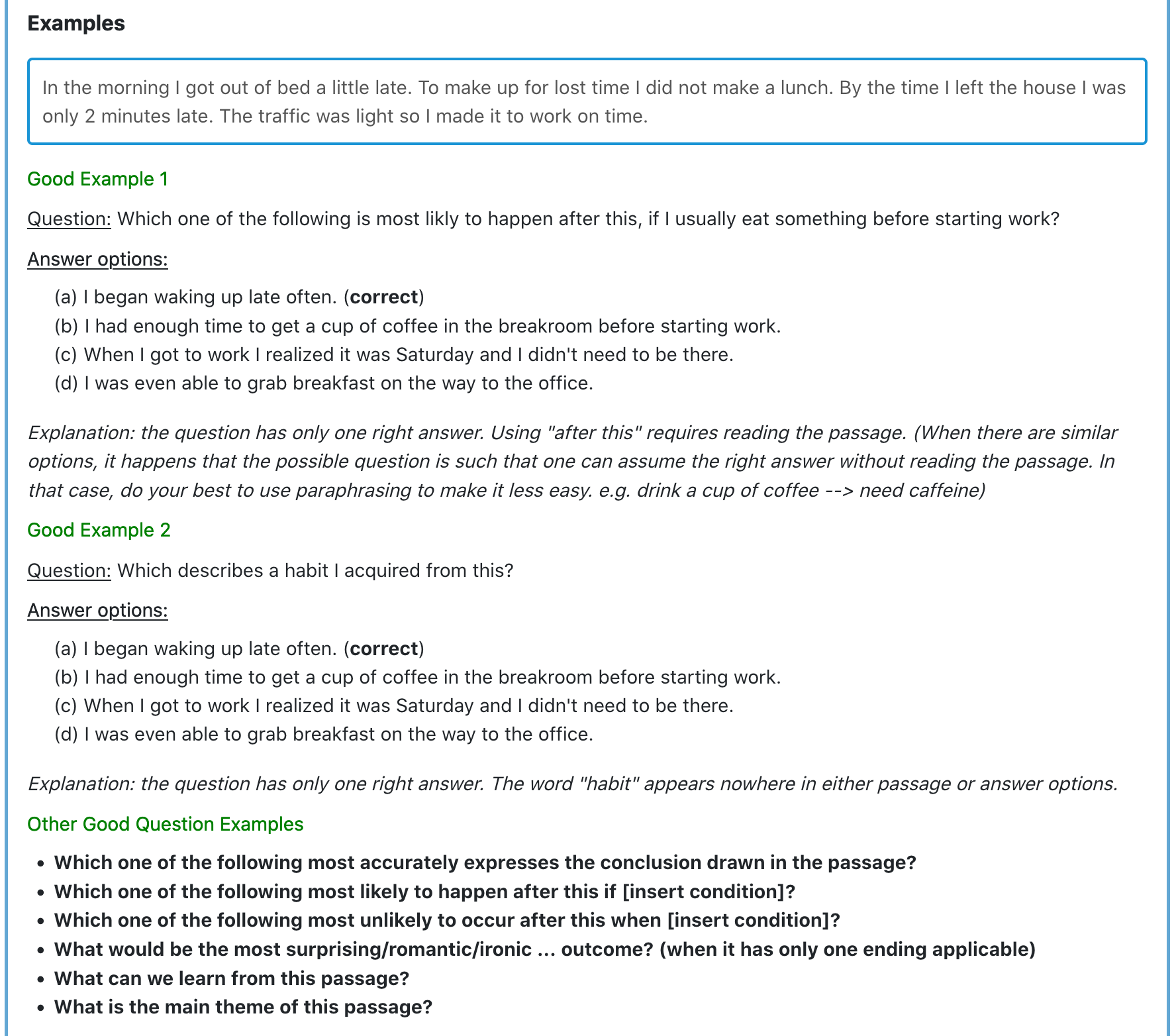}
    \caption{
        Instructions (2/3) used in the question writing task.
    }
    \label{fig:question-writing-instructions2}
\end{figure*}
\begin{figure*}[t]
    \includegraphics[width=\linewidth]{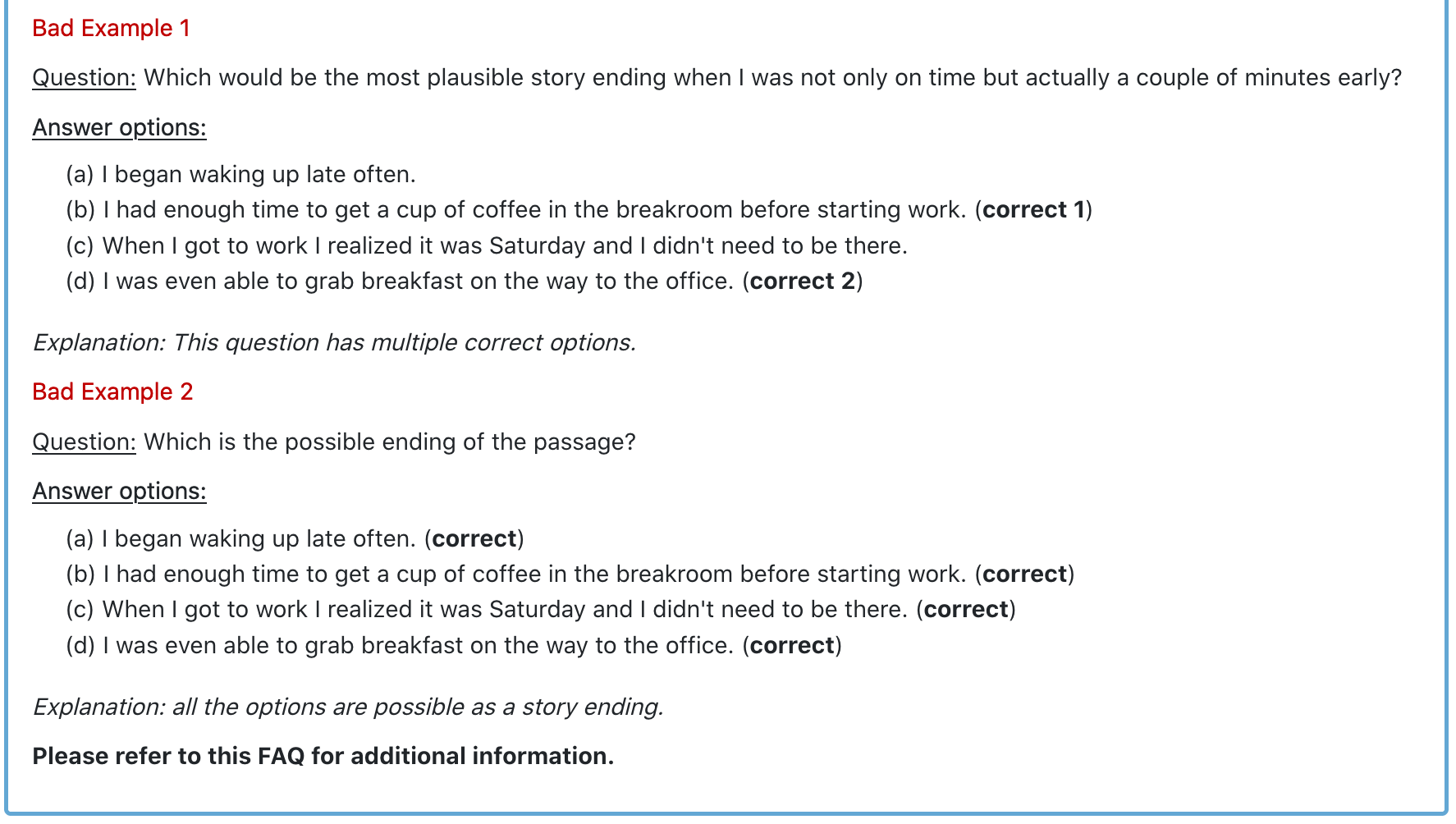}
    \caption{
        Instructions (3/3) used in the question writing task.
    }
    \label{fig:question-writing-instructions3}
\end{figure*}

\begin{figure*}[t]
    \includegraphics[width=\linewidth]{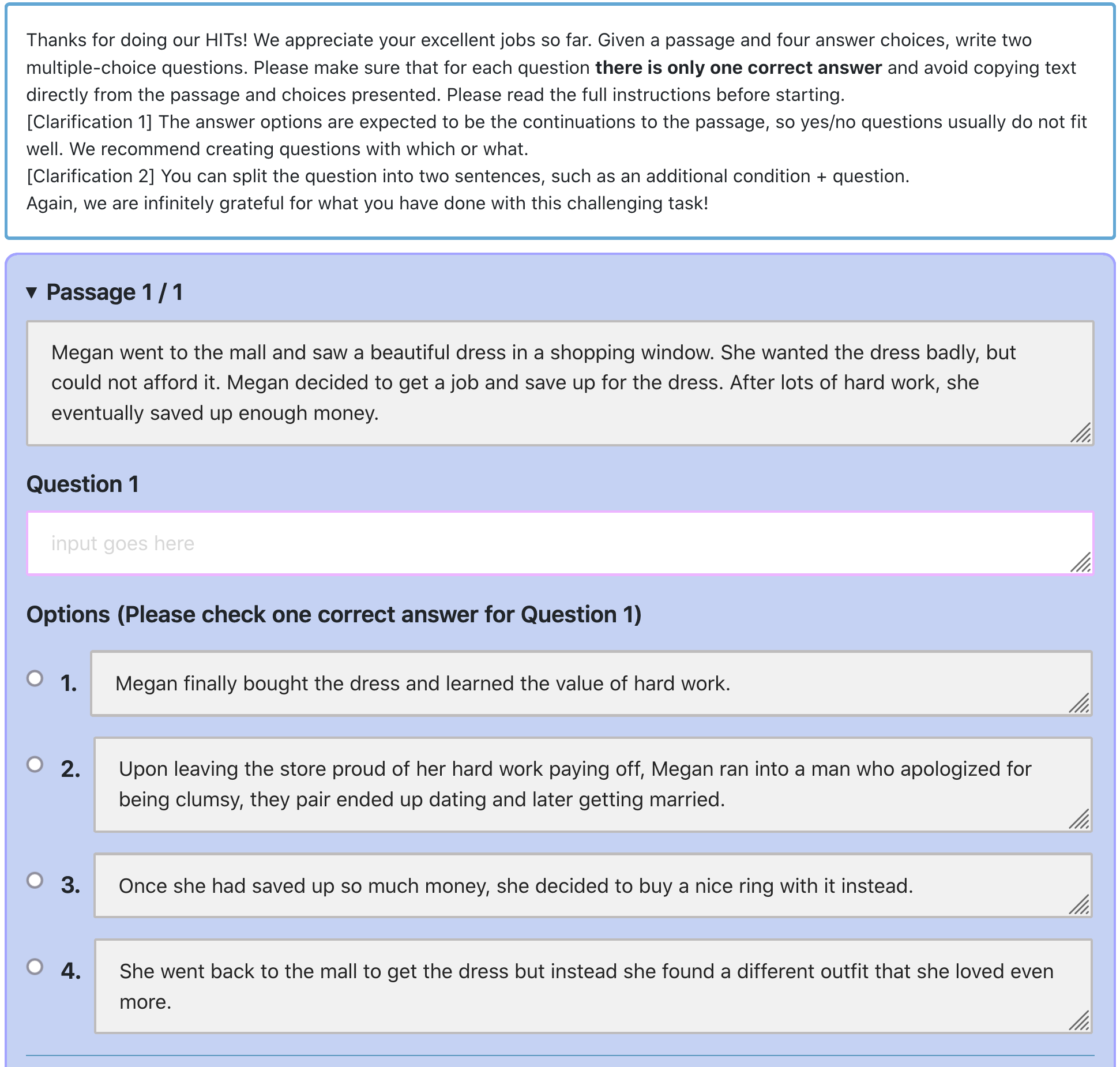}
    \caption{
        Interface (1/2) used in the question writing task.
    }
    \label{fig:question-writing-interface1}
\end{figure*}

\begin{figure*}[t]
    \includegraphics[width=\linewidth]{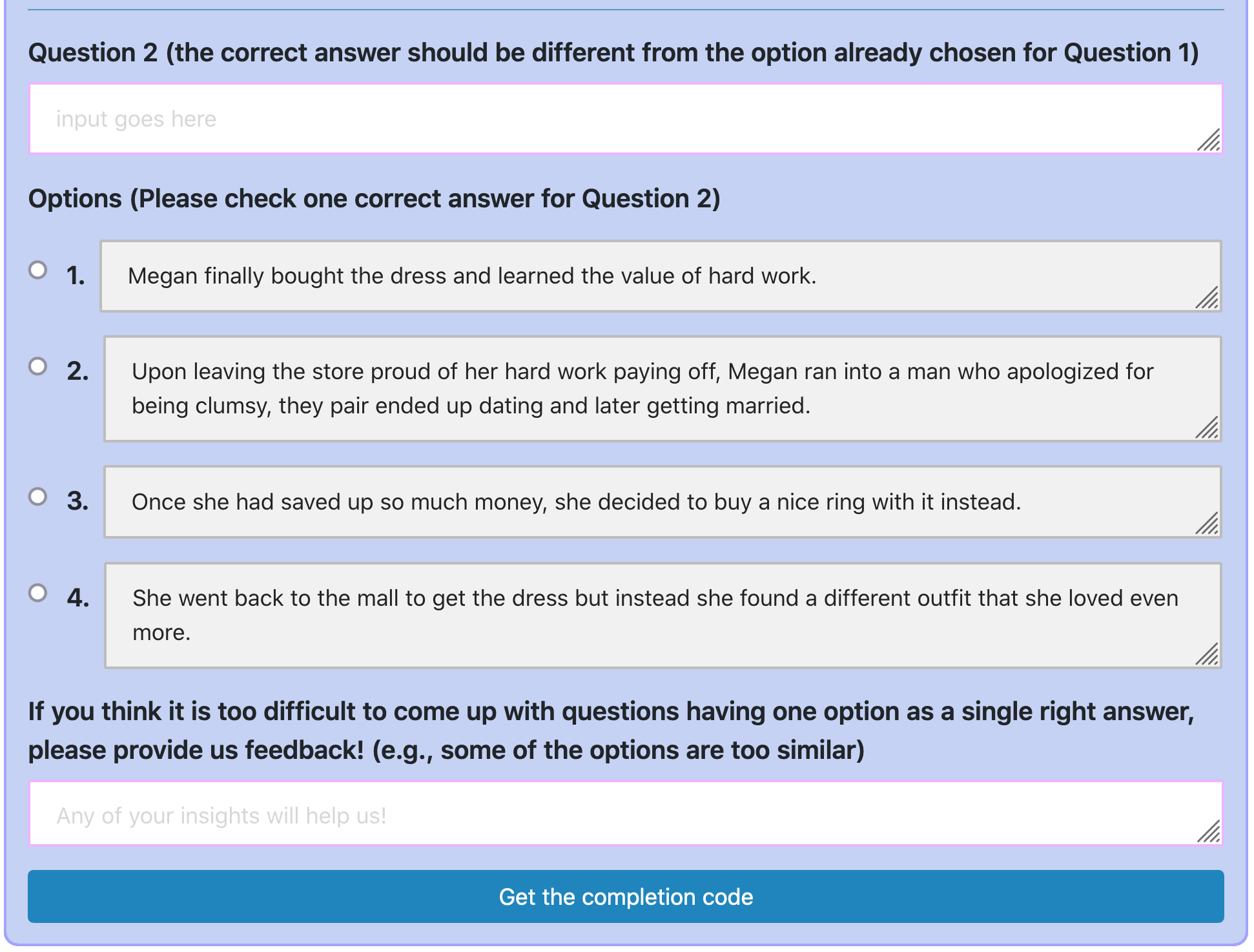}
    \caption{
        Interface (2/2) used in the question writing task.
    }
    \label{fig:question-writing-interface2}
\end{figure*}

Figures~\ref{fig:question-writing-instructions1}, \ref{fig:question-writing-instructions2}, and \ref{fig:question-writing-instructions3} show the instructions used in the question writing task.
Figures~\ref{fig:question-writing-interface1} and \ref{fig:question-writing-interface2} show the interface used in the question writing task.

\subsection{Question Validation}
\label{app:annotation-instructions-validation}

\begin{figure*}[t]
    \includegraphics[width=\linewidth]{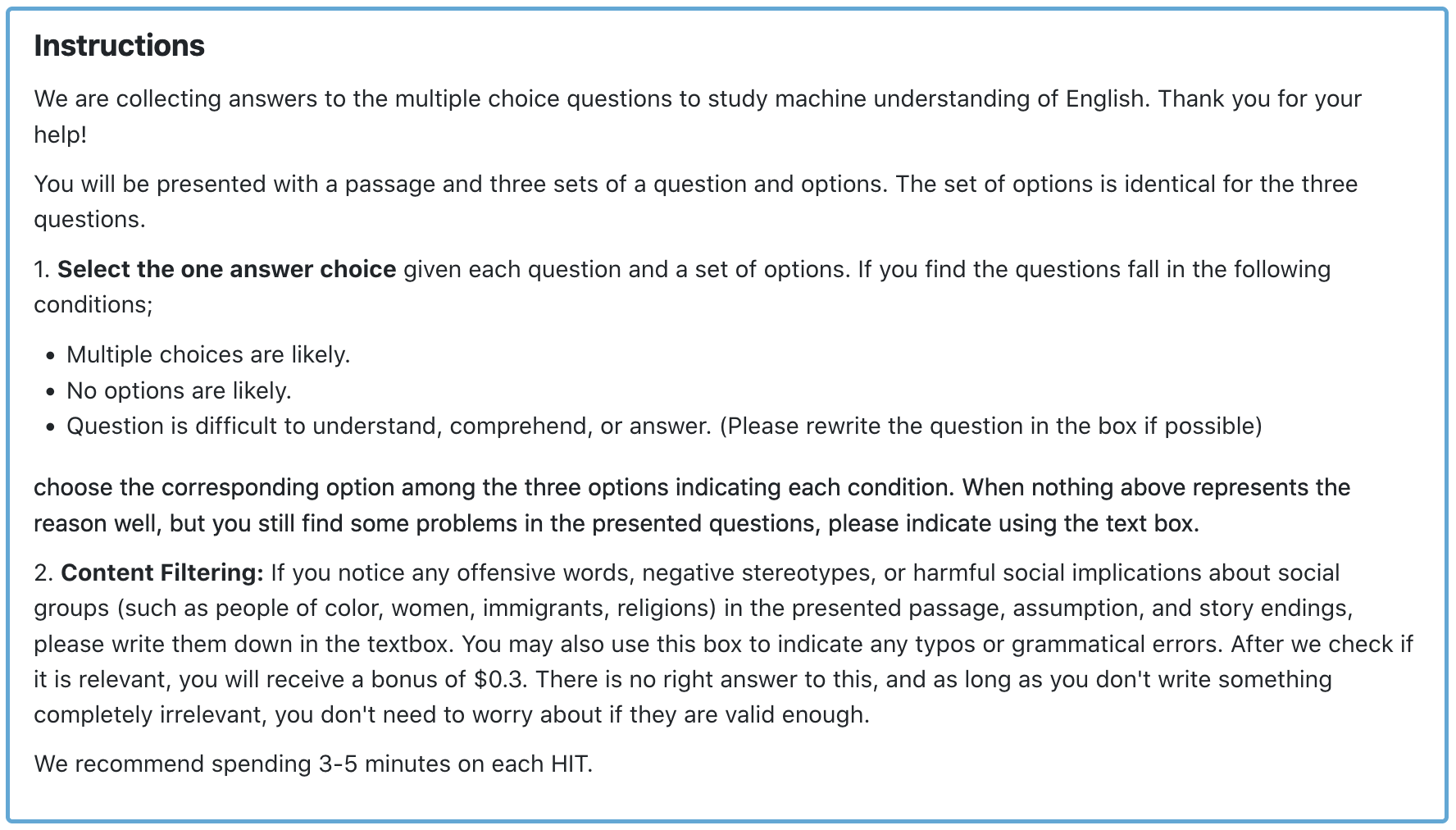}
    \caption{
        Instructions used in the question validation task.
    }
    \label{fig:validation-instructions}
\end{figure*}

Figure~\ref{fig:validation-instructions} shows the instructions used in the question validation task.

\end{document}